\documentclass[final, journal, doublecolumn, a4paper]{IEEEtran}
\usepackage{epsf,epsfig,amsmath,amssymb,subfigure,algorithm,algorithmic,url}

\usepackage{subfigure}

\usepackage{amsmath}
\usepackage{amsfonts}
\usepackage{amssymb}

\usepackage{algorithmic}
\usepackage{algorithm}

\usepackage{graphicx}
\usepackage{cite}
\usepackage{color}

\newcommand{\trans}[1]{{#1}^{\ensuremath{\mathsf{T}}}} 
\begin{document}

\title{Learning sparse representations of depth\thanks{This work has been supported by the Swiss National Science Foundation under the fellowship no:PBELP2-127847 awarded to I. To\v{s}i\'c.}}
\author{Ivana To\v{s}i\'c$^{\dag}$, Bruno A. Olshausen$^{\dag}$$^{\ddag}$ and Benjamin J. Culpepper$^{\ast}$\\
\dag Helen Wills Neuroscience Institute, University of California, Berkeley \\
\ddag School of Optometry, University of California, Berkeley\\
$\ast$ Department of EECS, Computer Science Division, University of California, Berkeley\\
ivana@berkeley.edu, baolshausen@berkeley.edu, bjc@cs.berkeley.edu}

\maketitle

\begin{abstract}
This paper introduces a new method for learning and inferring sparse representations of depth (disparity) maps. The proposed algorithm relaxes the usual assumption of the stationary noise model in sparse coding. This enables learning from data corrupted with spatially varying noise or uncertainty, typically obtained by laser range scanners or structured light depth cameras. Sparse representations are learned from the Middlebury database disparity maps and then exploited in a two-layer graphical model for inferring depth from stereo, by including a sparsity prior on the learned features. Since they capture higher-order dependencies in the depth structure, these priors can complement smoothness priors commonly used in depth inference based on Markov Random Field (MRF) models. Inference on the proposed graph is achieved using an alternating iterative optimization technique, where the first layer is solved using an existing MRF-based stereo matching algorithm, then held fixed as the second layer is solved using the proposed non-stationary sparse coding algorithm. This leads to a general method for improving solutions of state of the art MRF-based depth estimation algorithms. Our experimental results first show that depth inference using learned representations leads to state of the art denoising of depth maps obtained from laser range scanners and a time of flight camera. Furthermore, we show that adding sparse priors improves the results of two depth estimation methods: the classical graph cut algorithm~\cite{Boykov:2001p4748} and the more recent algorithm of Woodford et al.~\cite{Woodford09}. 
\end{abstract}


\section{Introduction}

Finding efficient representations of depth or disparity maps is important for applications involving inverse problems such as depth denoising and inpainting (for example in view synthesis~\cite{view}), and depth map compression (e.g., in 3DTV~\cite{Kubota:2007p1352}). Inverse problems have been extensively studied for natural images, for example using wavelet representations~\cite{Chang00adaptivewavelet}. However, because of differences between image and depth statistics, it is not obvious that wavelets are the most efficient way to represent the structure of depth maps. Thus, we prefer to \emph{learn} an efficient representation from a large database of examples. Sparse coding~\cite{olshausen97sparse} uses this approach to find overcomplete dictionaries of waveforms (atoms) in which the data has a sparse decomposition.
Sparse coding and other dictionary learning techniques have been successfully applied to learning image~\cite{olshausen97sparse, Engan:focus,Lewicki:2000p2757,Aharon:2006p284, mairal:214,Yaghoobi09} and audio~\cite{5332363} representations. Imposing sparse, non-Gaussian priors over latent variables in a linear generative model leads to a learning rule which produces dictionaries with elements that capture non-trivial aspects of the data statistics, such as long-range spatial correlations. These learned dictionaries, which capture higher-order dependencies in the data, can be used to regularize methods for solving inverse problems, yeilding state-of-the art performance in denoising~\cite{Aharon06}.

Most algorithms for dictionary learning assume that the signal is corrupted by stationary additive white Gaussian noise. While this is often a valid assumption in natural images, it does not hold for depth data. Even when measuring depth directly with range scanners, noise varies locally due to the different reflection of scanner light pulses around transparent or reflective surfaces, or near boundaries. Likewise, estimation of disparity from stereo images using standard computer vision algorithms yields disparity maps with variable uncertainty at each pixel in the map. Therefore, learning representations of depth requires adaptation of learning algorithms in order to deal with non-stationary noise in depth maps or with the unreliability of disparity map estimates. One contribution of this paper is a new learning algorithm based on sparse coding that is able to cope with non-stationary depth estimation errors. Noise statistics are inferred along with sparse coefficients during the inference step, which are then passed to the learning step that properly incorporates this uncertainty into the adaptation of the dictionary. This allows the dictionary learning method to be spatially adaptive and robust to noise. We show that this sparse coding method gives state-of-the-art performance in denoising of depth maps.

Learned representations of disparity are also important priors in depth estimation from stereo images, which is still a challenging problem in computer vision and robotics. Our second contribution is a new stereo matching algorithm that exploits the sparse prior over the learned depth atoms, which allows for modeling higher-order dependencies in depth map data. Such higher order priors encompass more information about the 3D structure than smoothness priors typically used in computer vision. We define stereo matching as a maximum aposteriori depth estimation problem on a two layer graphical model, where the top layer consists of hidden units that represent the coefficients in the sparse code of the depth map. The middle layer incorporates the output of the upper layer into a Markov Random Field (MRF) with neighborhood smoothness constraints. The probabilities for each depth estimate given by the sparse priors are used to refine the input to the MRF that can be defined and solved using any of the existing MRF-based stereo matching algorithms. Therefore, the proposed approach represents a generic way to include higher order priors to existing MRF-based algorithms in order to improve their solutions. Our final experiments on Tsukuba, Cones and Teddy datasets~\cite{middlebury} demonstrate that sparse priors can be used to regularize depth map estimates and quantitatively improve stereo matching results of the standard graph-cut algorithm (GC)~\cite{Boykov:2001p4748} and the more recent second order prior algorithm (2OP)~\cite{Woodford09}. 

%This is a challenging problem because of the ill-posed nature of the stereo correspondence problem. Namely, correlated features in the left and right images do not necessarily correspond to the same object in the 3D scene. Discovering methods that distinguish between false and correct stereo matches is of significant importance in both computational neuroscience and computer vision. Such a mechanism would allow us to understand binocular vision and furthermore to design systems that can infer the 3D world from 2D visual information. 

The paper is structured as follows. In Section~\ref{subsec:ns-SC}, we formulate the new sparse coding method using a generative model with non-stationary noise, and we present its energy minimization solution in Section~\ref{sec:EM}. Section~\ref{sec:stereo} describes depth inference from stereo based on sparse priors over learned disparity dictionaries. Experimental results in depth learning, denoising and inference from stereo are presented in Section~\ref{sec:results}. 
\section{Sparse coding with non-stationary additive Gaussian noise}\label{subsec:ns-SC}

The main principle underlying sparse coding (also called dictionary learning) theory is that some signals of dimension $N$ are well represented by a linear combination of a small number of elements selected from an overcomplete dictionary $\mathcal{D}$ of size $K>N$. This principle can be captured by the formalism of a linear generative model $\mathbf{f}=\mathbf{\Phi} \mathbf{a} + \boldsymbol{\epsilon}$. Columns of the matrix $\mathbf{\Phi}$ represent atoms from $\mathcal{D}$,  $\mathbf{a}$ is the vector of coefficients that weights each atom, and $\boldsymbol{\epsilon}$ is an $N$-dimensional vector of i.i.d. Gaussian noise of mean zero and variance $\sigma_0^2$. Since the dictionary is not given a priori, the main challenge of sparse coding is to learn the atoms in the dictionary given a set of training signals. Most approaches for dictionary learning are based on maximum likelihood estimation, where we look for the $\mathbf{\Phi}$ that maximizes $P(\mathbf{f}|\mathbf{\Phi})$~\cite{olshausen97sparse}, or the maximum aposteriori solution that maximizes $P(\mathbf{\Phi}| \mathbf{f})$~\cite{KreutzDelgado:2003p2758}.

Previous approaches for sparse coding assume that the additive noise $\boldsymbol{\epsilon}$ represents the portion of the signal that cannot be accounted for by the model, and it is usually modeled by an i.i.d. Gaussian noise process. However, depth maps acquired using laser range scanners and structured light contain noise that has spatially varying statistics. To account for this type of noise, we propose the following generative model for the depth map $\mathbf{f}$:
\begin{equation}
\label{eq:im_model2}
\mathbf{f}=\mathbf{\Phi} \mathbf{a} +\boldsymbol{\epsilon}+\boldsymbol{\eta},
\end{equation}
where $\boldsymbol{\eta}$ captures spatially varying noise from the sensor. We assume that this noise has a multivariate Gaussian distribution of zero mean and covariance matrix $\mathbf{\Sigma}_{\boldsymbol{\eta}}$, i.e., $ \boldsymbol{\eta}  \sim \mathcal{N}_N (0, \mathbf{\Sigma}_{\boldsymbol{\eta}})$. Since we have a sum of two Gaussian noises, the total noise $\zeta=\boldsymbol{\epsilon}+\boldsymbol{\eta}$ also has a Gaussian distribution $\mathcal{N}_N (0, \mathbf{\Sigma})$, where $\mathbf{\Sigma}=\mathbf{\Sigma}_{\boldsymbol{\eta}}+\sigma_0^2 \mathbf{I}$.
This covariance matrix $\mathbf{\Sigma}$ represents a set of variables that we need to infer along with the coefficients $\mathbf{a}$. In the case of sparse coefficient vectors $\mathbf{a}$, our optimization problem is:
\begin{align}
\label{eq:opt_prob}
\mathbf{\Phi}^*&= \arg \max_{\mathbf{\Phi}}{P(\mathbf{f}|\mathbf{\Phi})}\nonumber \\
& \approx \arg\max_{\mathbf{\Phi}}\left[\max_{\mathbf{\mathbf{a}}, \mathbf{\Sigma}}{P(\mathbf{\Phi}|\mathbf{f}, \mathbf{\mathbf{a}},\mathbf{\Sigma}) P(\mathbf{\mathbf{a}}) P(\mathbf{\Sigma})}\right] .
\end{align}
In the rest of the paper, we consider only independent (and thus uncorrelated) external noise, such that $\mathbf{\Sigma}_{\boldsymbol{\eta}}=\text{diag}({\tilde{\sigma}_1^2,...,\tilde{\sigma}_N^2})$, and therefore we have: $\mathbf{\Sigma}=\text{diag}({\sigma_1^2,...,\sigma_N^2})$, where $\sigma_i^2=\sigma_0^2+\tilde{\sigma}_i^2$, for $i=1,...,N$. The case of correlated noise is also interesting, but outside the scope of this paper.

The conditional probability of $\mathbf{f}$, given $\mathbf{a}$, $\mathbf{\Phi}$ and $\mathbf{\Sigma}$ in this case is:
\begin{align}
\label{eq:noiseprob}
P(\mathbf{f}|\mathbf{\Phi},\mathbf{a},\mathbf{\Sigma})&=\frac{\exp{\left[-\frac{1}{2} \trans{(\mathbf{f}-\mathbf{\Phi} \mathbf{a})} \mathbf{\Sigma}^{-1} (\mathbf{f}-\mathbf{\Phi} \mathbf{a})\right]}}{(2\pi)^{N/2} |\mathbf{\Sigma}|^{1/2}} \nonumber \\
&=\prod_{i=1}^N \left[ \frac{1}{\sqrt{2\pi}|\sigma_i|} \exp{(-\frac{(f_i-\hat{f}_i)^2}{2\sigma_i^2} )} \right],
\end{align}
where $f_i$ and $\hat{f}_i$ are $i$-th entries of vectors $\mathbf{f}$ and $\mathbf{\hat{f}}=\mathbf{\Phi}\mathbf{a}$, respectively. For the prior on the coefficient vector $\mathbf{a}$ we take a Laplace distribution, which is peaked at zero and heavy tailed. This is a usual choice in most sparse coding methods. Therefore, we have: $P(\mathbf{a})\propto \exp{(-\lambda \|\mathbf{a}\|_1)}$, where $\lambda$ controls the sparsity of $\mathbf{a}$. 

Unlike in previous dictionary learning methods, we impose a hyperprior on the covariance matrix of the noise in our model. Because we assume the noise at each location is i.i.d., our hyperprior $P(\mathbf{\Sigma})$ is factorial, i.e., $P(\mathbf{\Sigma})=\prod_{i=1}^N P(\sigma_i)$. We do not know what shape this noise hyperprior should have, so we choose the non-informative Jeffreys prior for the noise variance at each depth sample $i$. The Jeffreys prior on the variance of normal distribution is simply $\frac{1}{|\sigma_i|}$, which gives:
 \begin{equation}
\label{eq:noise_prior}
P(\mathbf{\Sigma})=\prod_{i=1}^N P(\sigma_i)=\prod_{i=1}^N \frac{1}{|\sigma_i|}.
\end{equation}

With the defined priors, our optimization problem in Eq.~\ref{eq:opt_prob}, cast as an energy minimization problem, becomes:
\begin{align}
\label{eq:opt_prob2}
\mathbf{\Phi}^*&=\arg \min_{\mathbf{\Phi}}{E(\mathbf{f}|\mathbf{\Phi},\mathbf{a})}\nonumber\\
&=\arg \min_{\mathbf{\Phi}}\left[\min_{\mathbf{\mathbf{a}}, \{\sigma_i\}}{\sum_{i=1}^N \left[ \log{\sigma_i^2} +\frac{(f_i-\hat{f}_i)^2}{2\sigma_i^2} \right] +}\lambda \|\mathbf{a}\|_1  \right] \, ,
\end{align}
where $E(\mathbf{f}|\mathbf{\Phi})=-\log{P(\mathbf{f}|\mathbf{\Phi},\mathbf{a})}$. The following section explains how to minimize this energy function.
 
\section{Inference and learning in the sparse coding model with non-stationary noise}\label{sec:EM}
Both inference and learning are accomplished by minimizing the negative log probability of the data under the model. This is done via an alternating optimization technique that can be viewed as a variational approximation to the E-M algorithm. At the beginning of each iteration, we select a depth map patch and the pixel-wise noise variances are initialized to fairly large values. The inference consists of two parts. The energy is first minimized with respect to the coefficients. Next, with these coefficients held fixed, we minimize the energy with respect to the pixel-wise noise variances. These two steps are alternated until convergence, which usually happens in only a few iterations. Finally, in the learning step, we compute the gradient of the energy function with respect to the dictionary, using the inferred coefficients and noise variances, and take a small step in that direction (learning). The details of this scheme are described below.

\subsection{Inference}

Our inference step differs from the usual convex optimization done in sparse coding since we need to infer the variances of the noise at each depth sample. In Section~\ref{subsec:ns-SC}, we have seen that the total noise has two components: 1) the approximation noise $\boldsymbol{\epsilon}$ with variance $\sigma_0^2$ that is equal for all depth samples, and 2) the external noise $\boldsymbol{\eta}$ with variance $\tilde{\sigma}_i^2$ that differs at each sample. In the inference step, we will assume that $\sigma_0$ is fixed and we optimize with respect to $\tilde{\sigma}_i$. Note that this will not significantly influence the obtained results since we can put a small value for $\sigma_0$ and all the noise variability will be shifted to $\tilde{\sigma}_i$. However, it will ensure that $\sigma_i \neq 0$ and that the solution is stable. The inference step is then:
\begin{align}
\label{eq:inference_uncorr}
(\mathbf{a},\{\tilde{\sigma}_i\})^*&= \arg \min_{\mathbf{a},\{\tilde{\sigma}_i\}} [ \sum_{i=1}^N \log{(\sigma_0^2+\tilde{\sigma}_i^2}) \nonumber \\
&+ \frac{1}{2} \sum_{i=1}^N \frac{[f_i-\sum_j a_j \boldsymbol{\phi}_j(i)]^2}{\sigma_0^2+\tilde{\sigma}_i^2}+ \lambda \|\mathbf{a} \|_1 ],
\end{align}
where $\boldsymbol{\phi}_j$, $j=1,...,K$ are atoms from $\mathcal{D}$. We perform this optimization in two alternating steps.
First, we fix a large value for all $\tilde{\sigma}_i$'s and optimize with respect to $\mathbf{a}$. This case is the regular $l2-l1$ optimization, where the $\tilde{\sigma}_i$'s may all be different constants.
In the second step, we fix $\mathbf{a}$ and use a closed form solution for hyperparameters $\tilde{\sigma}_i$'s:
\[ \tilde{\sigma}_i^2 = \left\{ \begin{array}{ll}
         \mbox{$0$} & \mbox{if $\frac{1}{2}(f_i-\hat{f}_i)^2 < \sigma_0^2$} \, ;\\
         \mbox{$\frac{1}{2}(f_i-\hat{f}_i)^2 - \sigma_0^2$} & \mbox{otherwise}.\end{array} \right. \]
where $\hat{f}_i=\sum_j a_j \boldsymbol{\phi}_j (i)$. Each step is guaranteed to descend the energy function. In practice, convergence is usually achieved in the first few iterations.

\subsection{Learning}
Learning is accomplished by taking a small step in the direction of the negative gradient of the energy function with respect to the dictionary $\mathbf{\Phi}$:
\begin{equation}
\label{eq:der_dico}
\frac{\partial E}{\partial \mathbf{\Phi}} = -\langle \mathbf{\Sigma}^{-1}(\mathbf{f}-\mathbf{\Phi}\mathbf{a}^*)\trans{\mathbf{a}^*} \rangle \,  .
\end{equation}
Since we select different depth map patches at each iteration, this effectively averages learning gradients drawn from the entire data set, denoted by $\langle \cdot \rangle$. The learning rule differs from that of standard sparse coding in that the dictionary update is weighted by the inverse of the noise covariance matrix. This results in adaptive dictionary updates, where observations with smaller noise variance (more reliable ones) have a higher influence on the learning than the observations with high noise variance (unreliable ones). 

In the rest of the paper, for convenience we will refer to our sparse coding method with non-stationary noise as ns-SC. It is important to note that ns-SC can be used not only for learning from depth maps, but also in the following general cases:
\begin{enumerate}
\item when we learn from data acquired by sensors that introduce non-stationary noise;
\item when we learn from inferred data, where each inferred variable has a certain reliability (e.g., learning in layered models).
\end{enumerate}
The second case is certainly a very important one, as there are many examples encountered in nature where we need to learn or infer the states of some hidden variables from other inferred variables. A relevant example is depth inference from stereo using sparse priors, which we propose in the next section.

\section{Stereo matching with sparse priors}\label{sec:stereo}

The dense depth estimation problem is usually formulated as a MAP estimation problem. Given the left and right images $ \mathbf{L}$ and $\mathbf{R}$ respectively, we want to estimate a disparity\footnote{Since there is a unique mapping from disparity values to depth values, inferring disparity is an equivalent problem to inferring depth.} map $\mathbf{f}$. In other words, to each pixel $i$ in one of the images (the one that we choose as a reference) we need to assign a certain disparity value $f_i$. We propose an approach that combines the Markov Random Field (MRF) formulation of depth inference, commonly used in computer vision, and inference using higher-order sparse priors. We briefly review the MRF approach and then describe the proposed method for depth inference using sparse priors.

\subsection{MRF approach to depth inference}
Most state of the art depth estimation approaches in computer vision formulate depth inference by the following optimization problem:
\begin{align}
\label{eq:depthMAP}
\mathbf{f}^*= \arg \max_{\mathbf{f}} P(\mathbf{f}| \mathbf{L}, \mathbf{R}) = \arg \max_{\mathbf{f}} P(\mathbf{L}, \mathbf{R} | \mathbf{f} ) P(\mathbf{f}),
\end{align}
where $P(\mathbf{L}, \mathbf{R} | \mathbf{f} )$ is the data likelihood, and $P(\mathbf{f})$ is the joint prior for disparity variables $f_i$. The likelihood term is usually modeled with a factorial Gaussian distribution:
\begin{align}
\label{eq:likelihood}
P(\mathbf{L}, \mathbf{R} | \mathbf{f} ) &\propto \prod_{i=1}^N \exp{\left[ - \frac{(L_i -R_{i+f_i})^2}{2\rho^2} \right]}\nonumber\\
&=\exp{\left[ -\sum_{i=1}^N \frac{D(f_i)}{2\rho^2} \right]},
\end{align}
where $L_i$ is the value of left image at pixel $i$, $R_{i+f_i}$ is the value of the right image at pixel $j$ displaced from pixel $i$ by $f_i$, and $\rho^2$ is the stationary noise variance. In computer vision, the disparity $f_i$ is usually one-dimensional since the stereo images are rectified. However, in general one can also consider two dimensional disparities. The function $D(f_i)=(L_i -R_{i+f_i})^2$ is usually called the data consistency term. 

When the depth map is modeled by a Markov Random Field (MRF), the prior over disparities can be expressed as: 
\begin{align}
\label{eq:MRF}
P(\mathbf{f}) &\propto \exp{\left[- \sum_{c \in \mathcal{C} } V_c(\mathbf{f})\right]}\nonumber \\
&=\exp{\left[- \sum_i {V_1(f_i)} -  \sum_{i,j \in \mathcal{N}_i} {V_2(f_i,f_j) - ...} \right]},
\end{align}
where $\mathcal{C}$ is a set of cliques, and the $V_c$ are clique potentials~\cite{MRF} of first, second and higher order. A particularly interesting case is when the cliques are at most of order two, such that the prior includes pairwise correlations between disparity variables at neighboring nodes. In this case, the disparity estimation problem is: 
\begin{align}
\label{eq:depthE}
\mathbf{f}^*&= \arg \min_{\mathbf{f}} E(\mathbf{f}| \mathbf{L}, \mathbf{R}) \nonumber \\
&= \arg \min_{\mathbf{f}} \left[ \sum_i{\frac{D(f_i)}{2\rho^2}}  + \sum_i {V_1(f_i)} +  \sum_{i,j \in \mathcal{N}_i} {V_2(f_i,f_j)} \right],
\end{align}
where $V_1$ and $V_2$ are first and second order cliques, which can be defined in a number of different ways depending on the task at hand. $\mathcal{N}_i$ denotes the neighborhood nodes of node $i$. Such energies can be efficiently minimized by graph cut~\cite{Boykov:2001p4748}, belief propagation~\cite{JianSun:2003p290}, log-cut~\cite{logcut}, etc. When the first order cliques $V_1$ are equal (no preferred disparity), the energy function in Eq.~(\ref{eq:depthE}) reduces to the one used in most computer vision algorithms. Second order cliques incorporate the smoothness constraint, and they can be evaluated as absolute distance between disparities of neighboring nodes, or by the Potts energy that puts more weight on neighboring that disparities differ. However, a pairwise model such as (\ref{eq:depthE}) cannot incorporate higher order structure such as depth edges. Although the MRF model can be extended to include triplewise correlations~\cite{Woodford09}, including even higher order priors in a single layer leads to high complexity graphs, which in general cannot be optimized by the graph cut~\cite{Kolmogorov04}. 

Another approach to modeling higher order dependencies is via a sparse prior over a dictionary adapted to the structure of signals. Such an approach has proven successful in natural images, where a sparsity prior on a dictionary of oriented edges solves inverse problems such as denoising and inpainting~\cite{olshausen97sparse, Lewicki99, Aharon06}. We expect that such priors would also play a crucial role in solving the correspondence problem, and thus we approach the problem of depth inference by including a sparsity prior on local depth features learned by the algorithm proposed in Section~\ref{subsec:ns-SC}. The details of our solution are described in the next section.

\subsection{Depth inference using sparse priors}\label{subsec:stereo}
\begin{figure}
\begin{center}
\includegraphics[width=0.4\textwidth]{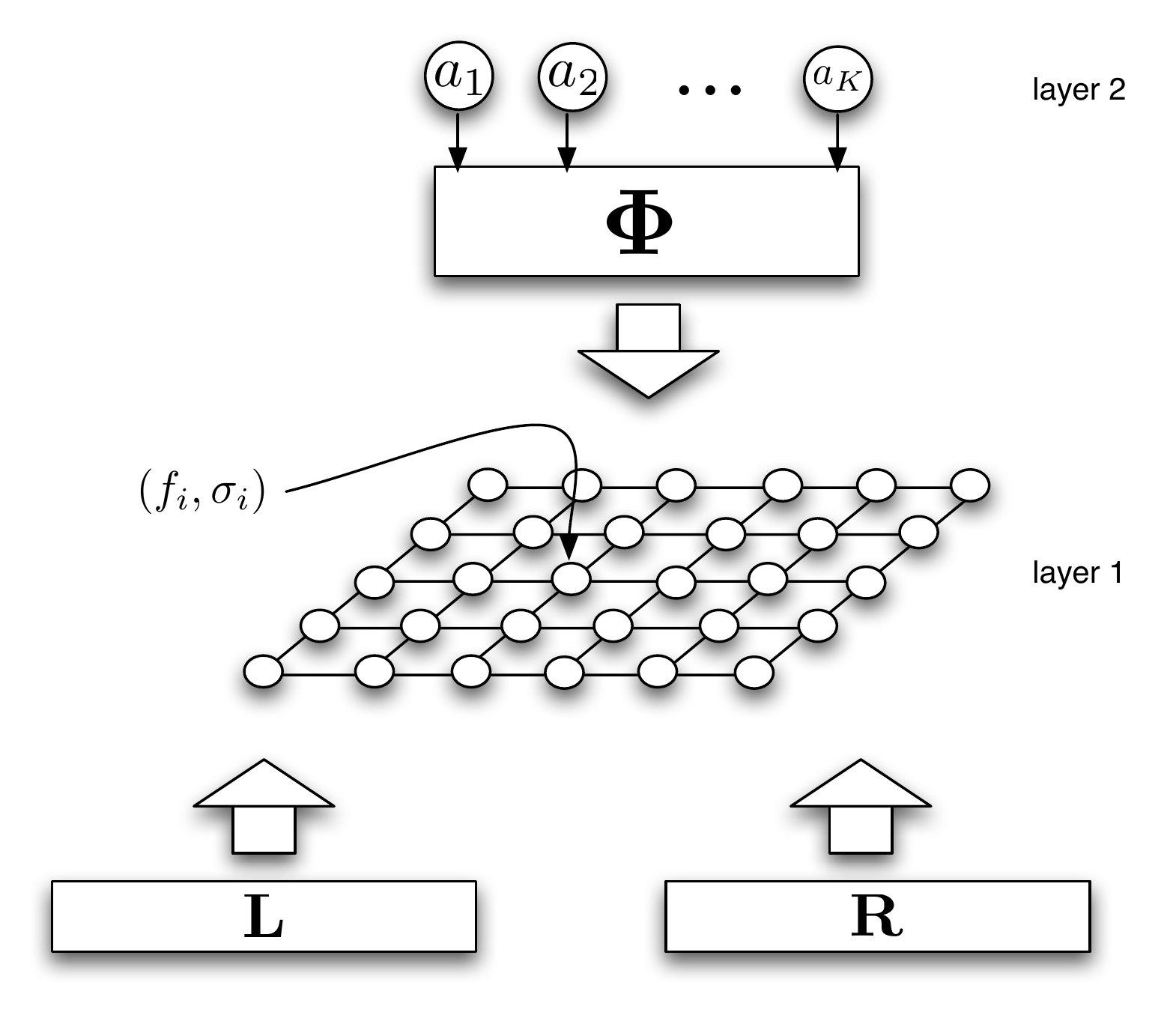}
\caption{Two-layer graphical model for depth inference.}
\label{fig:graph_model}
\end{center}
\end{figure}

We propose to combine the MRF structure with a sparse coding network within a two layer graphical model shown in Fig.~\ref{fig:graph_model}. The bottom layer is made up of the left and right input images. The middle layer is modeled as an MRF (as previously described), where each node consists of two latent variables: depth estimates $f_i$ and the reliability of each depth estimate given by $\sigma_i$. The top layer consists of latent sparse coefficients $a_j$ that capture higher order depth dependencies. The depth inference problem is cast as:
\begin{align}
\label{eq:depthMAPns}
\mathbf{f}^*&= \arg \max_{\mathbf{f}} P(\mathbf{f}| \mathbf{L}, \mathbf{R}, \mathbf{\Sigma}, \mathbf{a})\nonumber\\ 
&= \arg \max_{\mathbf{f}} P(\mathbf{L}, \mathbf{R} | \mathbf{f},\mathbf{\Sigma}, \mathbf{a}) P(\mathbf{f}|\mathbf{\Sigma}, \mathbf{a}) P(\mathbf{\Sigma}) P(\mathbf{a}),
\end{align}
which has two additional variables with respect to Eq.~\ref{eq:depthMAP}: the covariance matrix $\mathbf{\Sigma}$ of depth noise (i.e., the variance or reliability of each depth estimate $\sigma_i$), and the sparse coefficients $\mathbf{a}$ that represent hidden units. We propose to solve this problem by alternating inference in each of the layers separately, i.e., the algorithm iterates between the estimation of $\mathbf{f}$ in the middle MRF layer and the inference of $\mathbf{a}$ and $\mathbf{\Sigma}$ by the sparse priors in the top layer. 

In inferring the middle layer, $\mathbf{a}$ and $\mathbf{\Sigma}$ are fixed and $\mathbf{f}$ is inferred as:
\begin{align}
\label{eq:depthEns}
\mathbf{f}^*&= \arg \min_{\mathbf{f}} E(\mathbf{f}| \mathbf{L}, \mathbf{R}, \mathbf{\Sigma}, \mathbf{a})\nonumber\\
& = \arg \min_{\mathbf{f}}[ \sum_i{\frac{D(f_i)}{2[\rho(\sigma_i)]^2}}  \nonumber\\
&+ \sum_i {\frac{(f_i-\hat{f}_i)^2}{2\sigma_{i}^2}} +  \sum_{i,j \in \mathcal{N}_i} {V_2(f_i,f_j)}],
\end{align}
where $\hat{f}_i$ is an element of the vector $\hat{\mathbf{f}}=\mathbf{\Phi} \mathbf{a}$. Since $\mathbf{a}$ and $\mathbf{\Sigma}$ are constant in this layer, priors $P(\mathbf{a})$ and $P(\mathbf{\Sigma})$ vanish from the energy function. This problem is similar to the inference problem in Eq.~\ref{eq:depthMAP}, with two differences: a) the data term variance $\rho$ depends on the variance of depth estimates $\sigma_i^2$ and is different for each $f_i$; and b) clique potentials depend on $\mathbf{a}$ and $\sigma_i^2$ and are given as $V_1(f_i) = (f_i-\hat{f}_i)^2/(2\sigma_{i}^2)$. The data term variances $[\rho(\sigma_i)]^2$ can be estimated by their expected values around $\hat{f}_i$ under the noise variance $\sigma_i$. We use a square window $(\hat{f}_i-\sigma_i, \hat{f}_i+\sigma_i)$ to calculate this expectation, i.e.,:
\begin{equation}
\label{eq:vardata}
[\rho(\sigma_i)]^2=\langle (D(f_i)-D(\hat{f}_i))^2 \rangle, \hspace{0.3cm} \forall {f_i \in (\hat{f}_i - \sigma_{i}, \hat{f}_i + \sigma_{i}) }.
\end{equation}

Once we have inferred $\mathbf{f}$ in the middle layer, the inference in the upper layer becomes:
\begin{align}
\label{eq:depthEns}
(\mathbf{a},\{\sigma_i \})^*&= \arg \min_{(\mathbf{a},\{\sigma_i \})} E(\mathbf{a}, \mathbf{\Sigma}| \mathbf{f})\nonumber\\ 
&= \arg \min_{(\mathbf{a},\{\sigma_i \})} \left[ \sum_i {\frac{(f_i-\hat{f}_i)^2}{2\sigma_{i}^2}} +  \| \mathbf{a} \|_1 \right],
\end{align}
where the hidden units are inferred by the ns-SC from the disparity estimates obtained by the middle layer. Since ns-SC evaluates both the mean $\hat{f}_i$ and the variance $\sigma_{i}^2$, it sends feedback to the middle layer to update the states of these variables. With new estimates for each variable $f_i$ and $\sigma_{i}^2$, the middle layer can re-evaluate the new disparity estimates with new clique potentials.

The main role of describing each node in the MRF by a Gaussian with mean $f_i$ and variance $\sigma_i^2$ is to resolve ambiguities in stereo matching. Namely, when the data likelihood at a certain point is unreliable ($\rho(\sigma_i)$ is large), the stereo matching algorithm puts more weight on the prior given by $V_1(f_i) $, which is estimated by the sparse priors from the upper layer. This happens, for example, at scene points with specular illumination, where the surface is not Lambertian. Since data consistency is violated at those points, we need to use prior information to solve the correspondence problem.

\section{Experimental results}\label{sec:results}

\subsection{Learning of depth dictionaries}\label{subsec:results_learning}

\begin{figure*}[!htbp]
\begin{center}
$\begin{array}{c@{\hspace{0.5 cm}}c}
\includegraphics[width=0.26\textwidth]{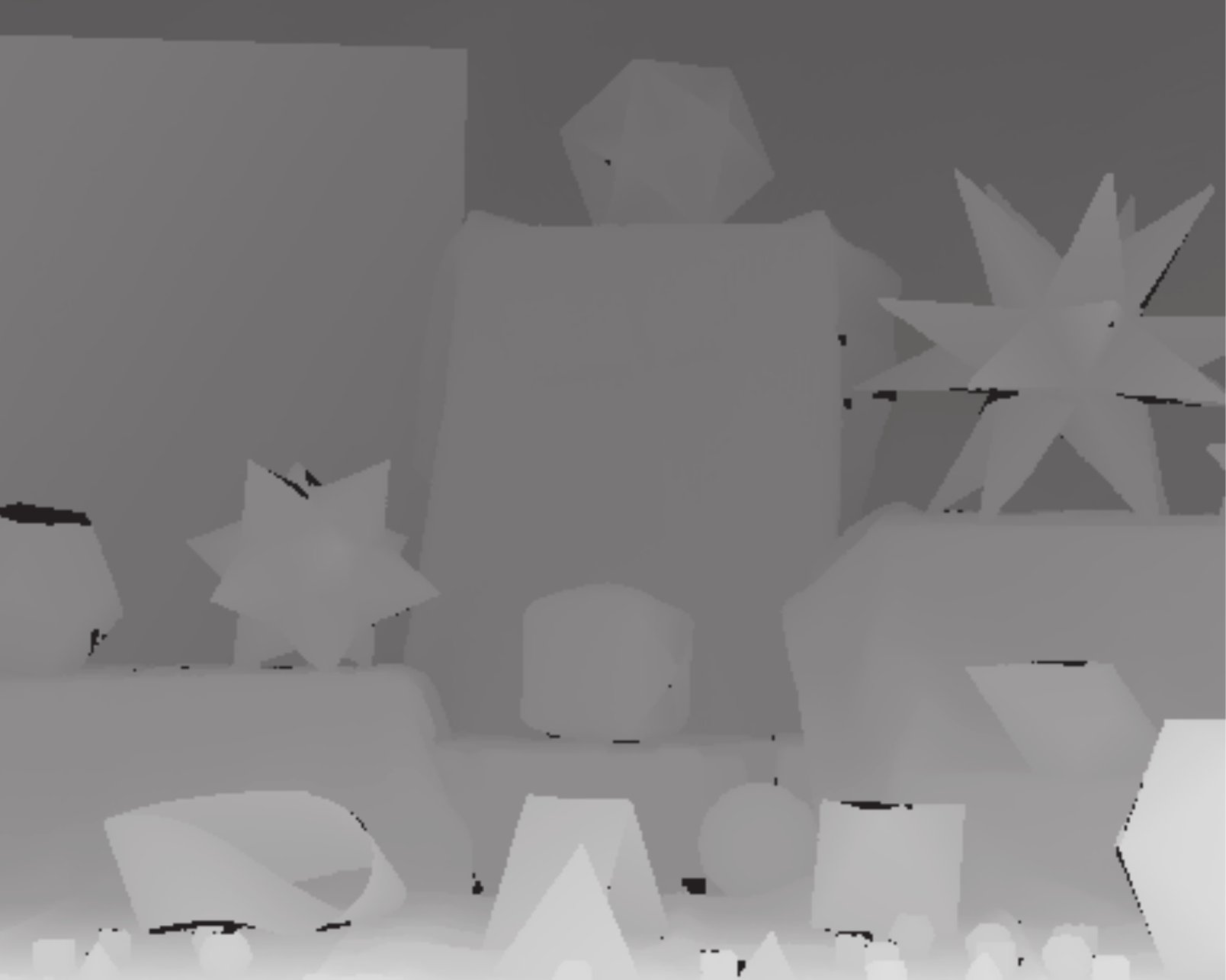}&
\includegraphics[width=0.25\textwidth]{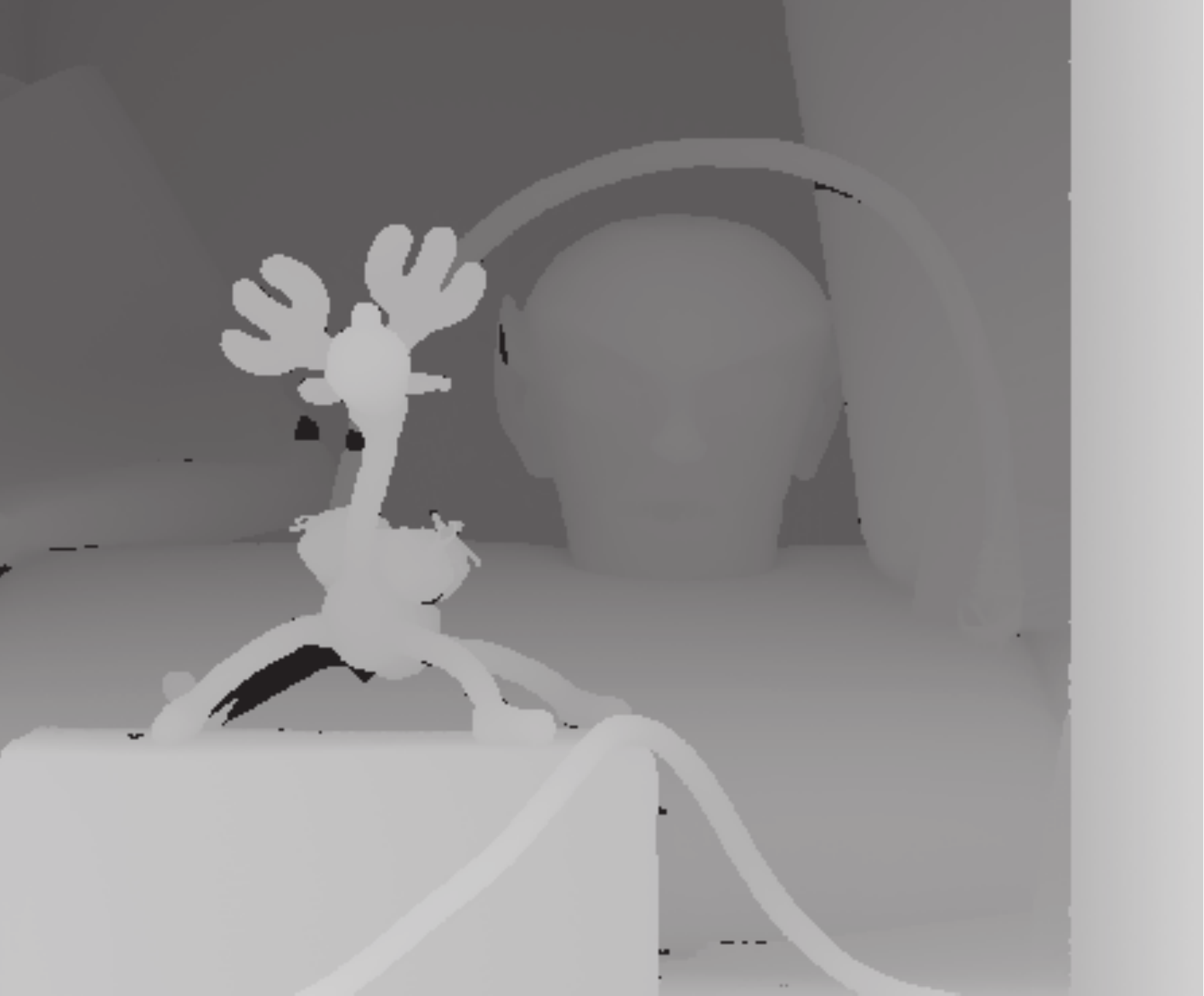} \\
\mbox{\bf (a) Moebius } & \mbox{\bf (b) Reindeer }\\
\end{array}$
\end{center}
\caption{Examples of disparity maps from the Middlebury 2006 dataset.}
\label{fig:maps}
\end{figure*}

We have learned overcomplete dictionaries of depth atoms using the regular sparse coding (SC)~\cite{olshausen97sparse} and ns-SC. For training, we have used the ground truth disparity maps from the 2005 and 2006 Middlebury stereo datasets~\cite{middlebury}. Two examples of disparity maps are shown in Fig.~\ref{fig:maps}. These maps represent ``inverse" depth, i.e., the disparity is inversely proportional to depth, but keeps the same features (e.g. edges) as depth. It also represents ``projective depth", because these disparity values are dependent on the viewing angle. Finding representations for the projective depth is especially desirable in multi-view (3DTV) technologies, where an image is usually aligned to a depth map in order to simplify view synthesis. On the other hand, laser range scanners are typically of different resolution and sampling than images, which makes them hard to register. Another possibility would be to learn on depth maps from time of flight cameras (TOF). However, there are no publicly available databases for TOF data. Therefore, we have chosen the Middlebury database for learning. No prior whitening has been performed.

Depth maps from the Middlebury stereo dataset were obtained using the structured light technique and they have missing pixels (black pixels in Fig.~\ref{fig:maps}). We treat those pixels in two ways:
\begin{enumerate}
  \item \textbf{Approach 1}: set the noise variance at missing pixels to infinity (their contribution to learning is thus zero), while the variance values of the rest of the pixels are inferred during ns-SC learning; and
  \item \textbf{Approach 2}: treat those pixels in the same manner as other pixels and perform ns-SC learning (i.e., variances of all pixels are inferred).
\end{enumerate}
The dictionary learned with Approach 1 is denoted as $\mathbf{\Phi}_1$ and with Approach 2 as $\mathbf{\Phi}_2$. The idea behind learning with Approach 2 is to see how well the ns-SC algorithm would do if it had no information where the missing pixels are, and would have to infer that. For comparison, we have also learned the dictionary using standard sparse coding with constant noise variance within the map~\cite{olshausen97sparse}, except at the missing pixels where the variance is infinity (i.e., missing pixels do not bias the learning). This dictionary is denoted as $\mathbf{\Phi}_3$.

\begin{figure*}[!htbp]
\begin{center}
$\begin{array}{c@{\hspace{0.5 cm}}c@{\hspace{0.5 cm}}c}
\includegraphics[width=0.3\textwidth]{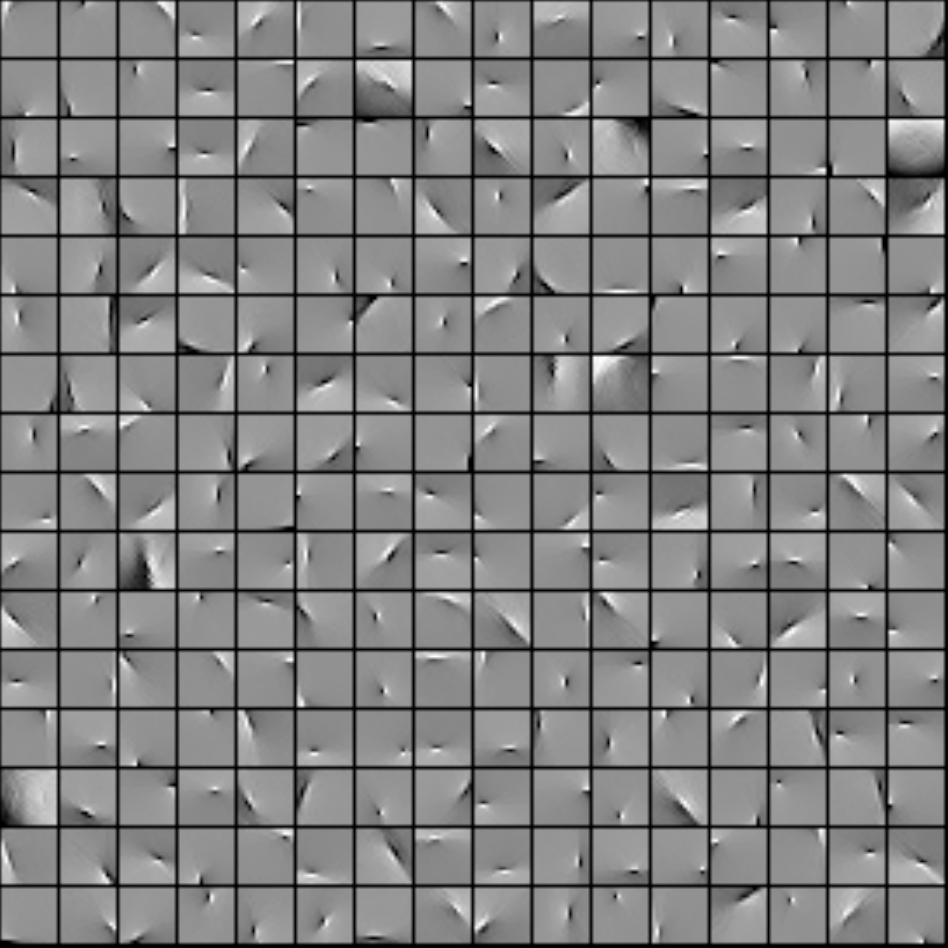}&
\includegraphics[width=0.3\textwidth]{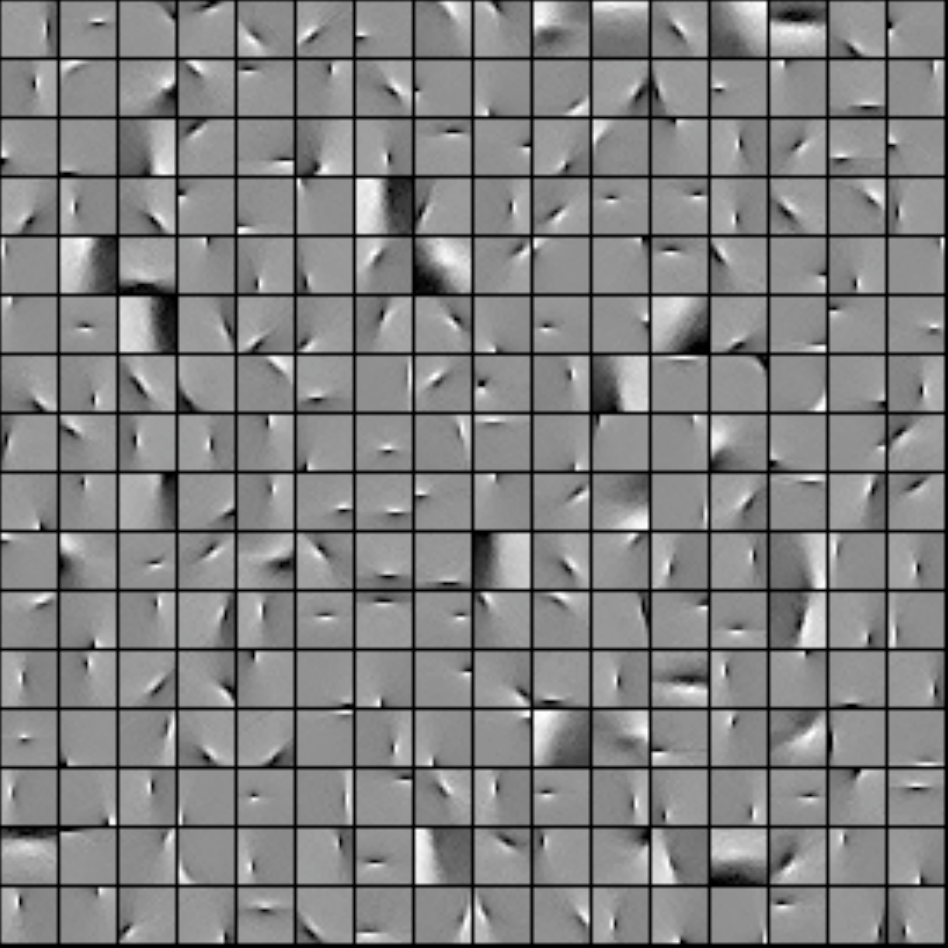}&
\includegraphics[width=0.3\textwidth]{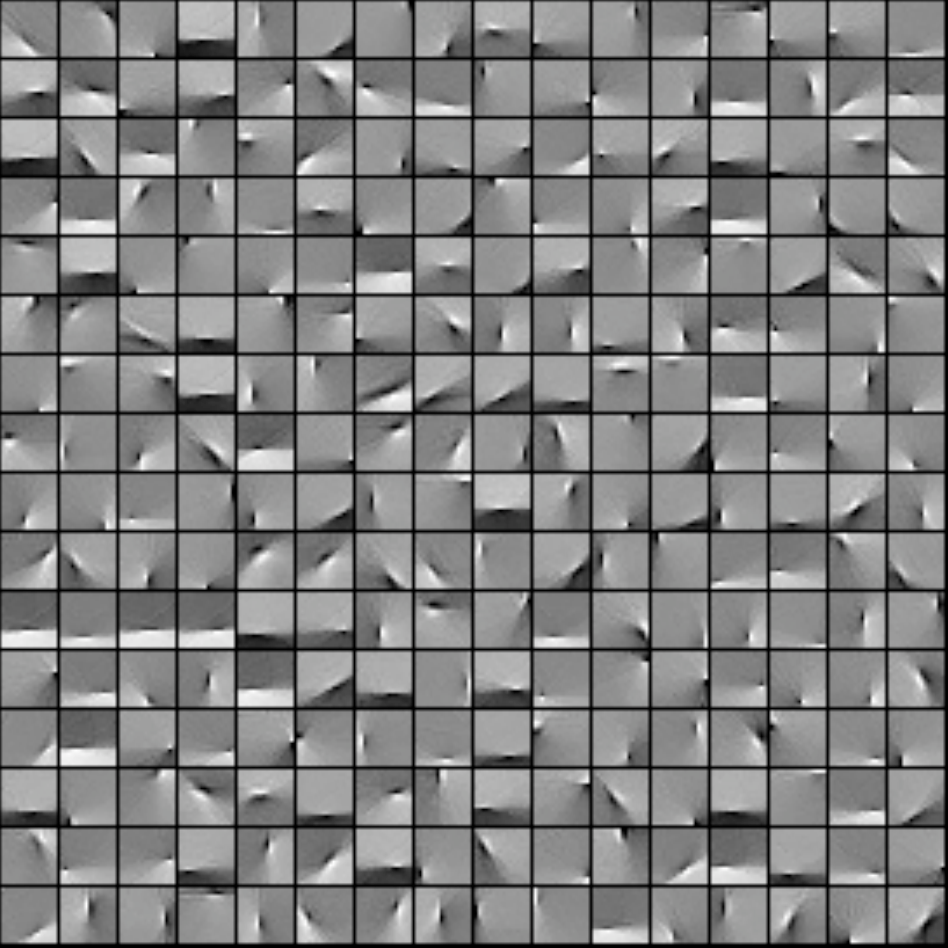} \\
\mbox{\bf (a) $\mathbf{\Phi}_1$ } & \mbox{\bf (b)$\mathbf{\Phi}_2$ }& \mbox{\bf (b)$\mathbf{\Phi}_3$ }\\\
\end{array}$
\end{center}
\caption{Learned depth dictionaries: a) with ns-SC and masked missing pixels (Approach 1); b) with ns-SC and no masking of missing pixels (Approach 2); c) with SC and masked missing pixels.}
\label{fig:visual2}
\end{figure*}

Within each iteration of ns-SC (and SC as well), we have randomly chosen a large set of $16\times 16$ patches. The dictionary size has been set equal to the signal size (256), in order to limit the complexity. Note that overcomplete dictionaries can be learned as well, leading to better performance in target applications and a higher computational cost. The parameter $\sigma_0^2$ has been chosen to have a small value $\sim 0.01$. Since the sparsity parameter $\lambda$ is subsumed by the variance of the noise inferred at each pixel, we set it to 1. Fig.~\ref{fig:visual2} displays learned dictionaries $\mathbf{\Phi}_1$, $\mathbf{\Phi}_2$ and $\mathbf{\Phi}_3$. We can see that $\mathbf{\Phi}_1$ and $\mathbf{\Phi}_2$ are qualitatively similar, with mostly edge-like depth functions and some slant-like atoms. These are the types of features usually seen in depth maps. The dictionary $\mathbf{\Phi}_3$ learned with SC exhibits some repetitive atoms, meaning that learning prefers some directions in the high-dimensional space. This might be explained by the fact that learning is done on unwhitened data, so some directions have higher energy (variance). This does not happen in ns-SC since the variance is inferred and the learning rule is adapted accordingly. However, it is hard to say which dictionary is the best based only on the qualitative assessment. Therefore, in the next section we perform quantitative comparison of these dictionaries on depth map denoising.

\subsection{Depth map denoising}
Depth maps obtained by laser range scanners and TOF cameras are typically corrupted by spatially varying noise. Denoising of these depth maps thus becomes an important step in applications that involve view synthesis in hybrid (depth plus video) camera systems~\cite{5413941}. Depth denoising can be achieved with the inference step of the ns-SC algorithm. For evaluation purpose we have added synthetic noise to 1\% of randomly chosen pixels in a depth map block of $100\times 100$ pixels, taken from the Purves natural depth maps database~\cite{Purves03} (different from the depth maps used for learning). The noise has been generated according to the non-stationary Gaussian model. Each pixel is corrupted by Gaussian noise whose variance is randomly chosen from 0 to 1. We divide a depth map into overlapping patches of $16\times16$ pixels (patches are shifted by 1) and denoise each patch as: $\mathbf{\hat{y}}=\mathbf{\Phi}\hat{\mathbf{a}}$, where $\hat{\mathbf{a}}$ is the sparse coefficient vector inferred by ns-SC. Prior to inference, each patch has been normalized to have a variance 1, which facilitates the choice of the initialization of noise variances (initialized also to 1). Each pixel in the denoised depth map is evaluated as an average over patches that overlap at that pixel. 

\begin{figure*}[!htbp]
\begin{center}
$\begin{array}{c@{\hspace{0.1 cm}}c@{\hspace{0.1 cm}}c@{\hspace{0.1 cm}}c}
\includegraphics[width=0.22\textwidth]{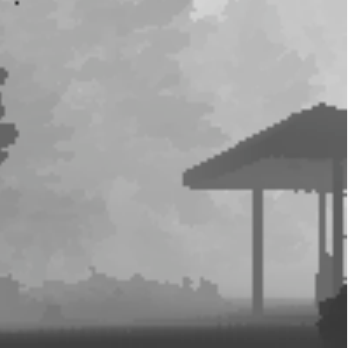}&
\includegraphics[width=0.22\textwidth]{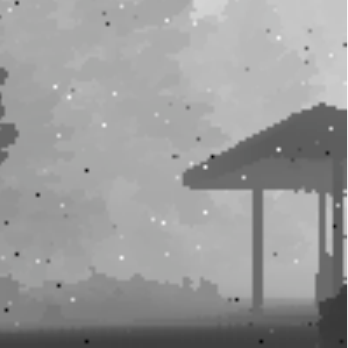} &
\includegraphics[width=0.22\textwidth]{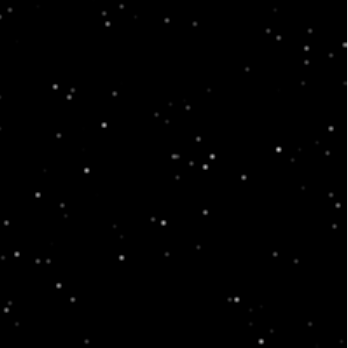} &
\includegraphics[width=0.22\textwidth]{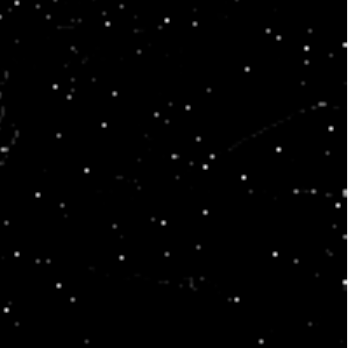}\\
\mbox{ (a) original} & \mbox{(b) noisy 28.5dB} & \mbox{(c) noise magnitude} & \mbox{(d) inferred variance}\\
\includegraphics[width=0.22\textwidth]{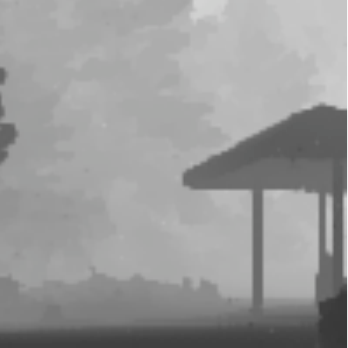}&
\includegraphics[width=0.22\textwidth]{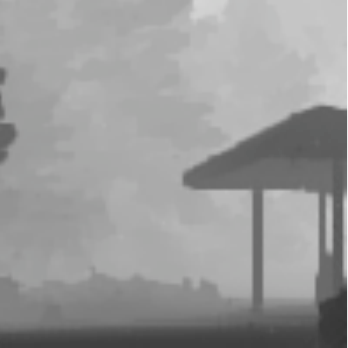}&
\includegraphics[width=0.22\textwidth]{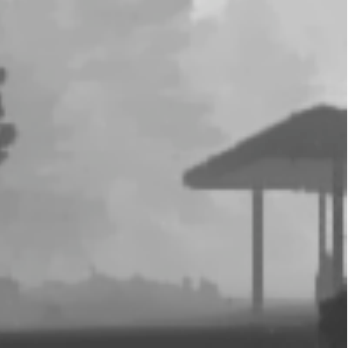}&
\includegraphics[width=0.22\textwidth]{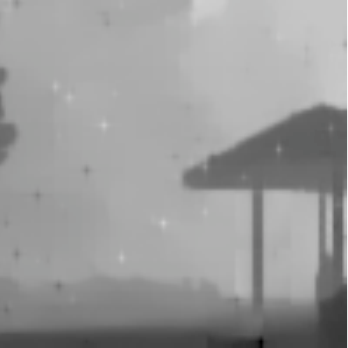}\\
\mbox{(e) ns-SC($\mathbf{\Phi}_1$), 37.8 dB} & \mbox{(f) ns-SC($\mathbf{\Phi}_2$), 36.3 dB} & \mbox{(g) ns-SC ($\mathbf{\Phi}_3$), 35.3 dB} & \mbox{ (h) SC, 32.3dB}\\
\includegraphics[width=0.22\textwidth]{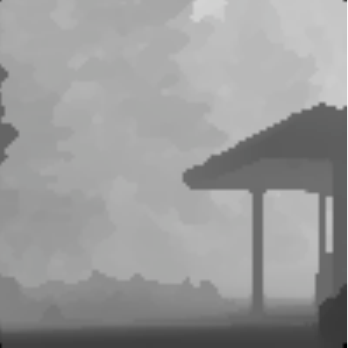}&
\includegraphics[width=0.22\textwidth]{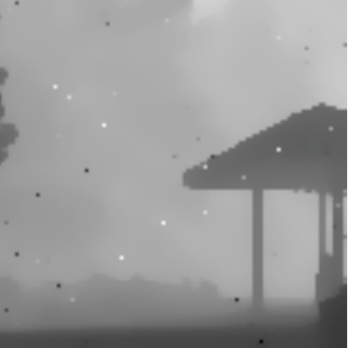}&
\includegraphics[width=0.22\textwidth]{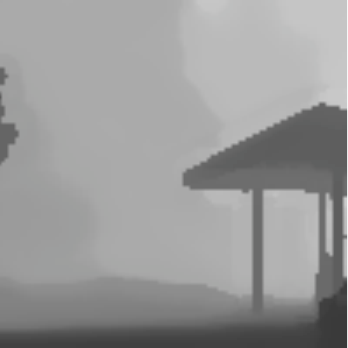}&
\includegraphics[width=0.22\textwidth]{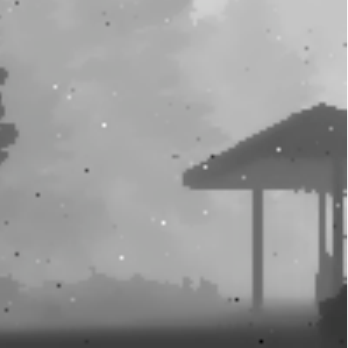}\\
\mbox{(i) median, 31.3 dB} & \mbox{(j) TV, 31.8 dB} & \mbox{(k) NL means, 32.9dB} & \mbox{(l) TIWF, 33.7dB}\\
\end{array}$
\end{center}
\caption{Denoising results for a depth map from the Purves database with synthetic noise (1\% of corrupted pixels). Performance of ns-SC using different dictionaries $\mathbf{\Phi}_1$, $\mathbf{\Phi}_2$ and $\mathbf{\Phi}_3$ and performance of other denoising methods: SC- sparse coding, TV- total variation, Median, Non-Local (NL) means and Translation Invariant Wavelet 7-9 Frame (TIWF). The variance inferred using the ns-SC method with $\mathbf{\Phi}_1$ is shown on subfigure (d).}
\label{fig:denoisingpurves}
\end{figure*}

The original and noisy depth maps are shown in Fig.~\ref{fig:denoisingpurves}a and Fig.~\ref{fig:denoisingpurves}b, respectively; while Fig.~\ref{fig:denoisingpurves}c shows the added noise. The denoised depth map using the ns-SC with $\mathbf{\Phi}_1$ (Fig.~\ref{fig:denoisingpurves}e) is of higher quality compared to the denoised depth map using ns-SC and $\mathbf{\Phi}_2$ (Fig.~\ref{fig:denoisingpurves}f) and ns-SC and $\mathbf{\Phi}_3$ (Fig.~\ref{fig:denoisingpurves}g), where the quality is measured by the Peak-SNR (PSNR). Besides denoising, ns-SC also estimates the variance of the noise, shown in Fig.~\ref{fig:denoisingpurves}d, whose spatial distribution (location of noisy pixels) corresponds to the noise pattern. Moreover, ns-SC using any one of the dictionaries outperforms denoising using classical SC with fixed variance $l2-l1$ minimization and $\mathbf{\Phi}_3$, shown in Fig.~\ref{fig:denoisingpurves}h. We have also performed median filtering denoising (Fig.~\ref{fig:denoisingpurves}i), since it is a proper filter for this type of noise; the Total Variation denoising~\cite{TVrof} using the algorithm of Chambolle~\cite{Chambolle04} (see Fig.~\ref{fig:denoisingpurves}j); the Non-Local (NL) means denoising using median filtering~\cite{buades:490} (Fig.~\ref{fig:denoisingpurves}k), and the ns-SC inference using the translation invariant wavelet 7-9 frame (TIWF, Fig.~\ref{fig:denoisingpurves}l). Again, ns-SC with any of the learned dictionaries $\mathbf{\Phi}_1$-$\mathbf{\Phi}_3$ outperforms the other solutions, both in PSNR and visual quality. Although the solutions obtained by median filtering and NL-means might also look visually pleasing, these types of filters average over the fine details in the map, which is undesirable. Using TIWF instead of the learned dictionaries for ns-SC (Fig.~\ref{fig:denoisingpurves}l) is also suboptimal since the wavelet frame is not adapted to the statistics of the signal. We do not report the comparisons with denoising methods designed for stationary noise (e.g., KSVD~\cite{Aharon06}, GSM~\cite{1240101} and BM3D~\cite{4271520}), since they are not really adapted to this type of noise and thus cannot appropriately handle it.

\begin{figure*}[!htbp]
\begin{center}
$\begin{array}{c@{\hspace{0.1 cm}}c@{\hspace{0.1 cm}}c}
\includegraphics[height=2in]{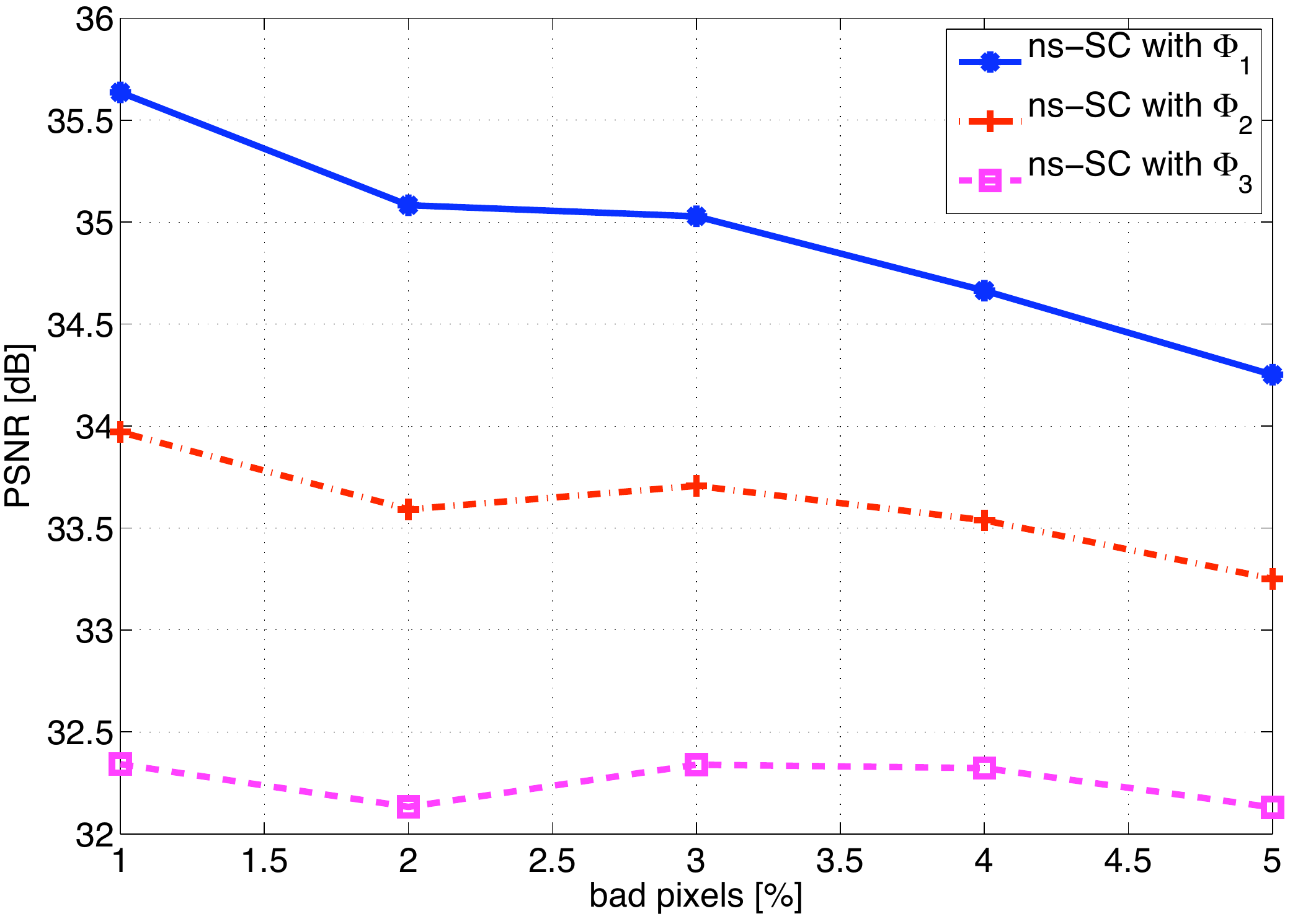}&
\includegraphics[height=2in]{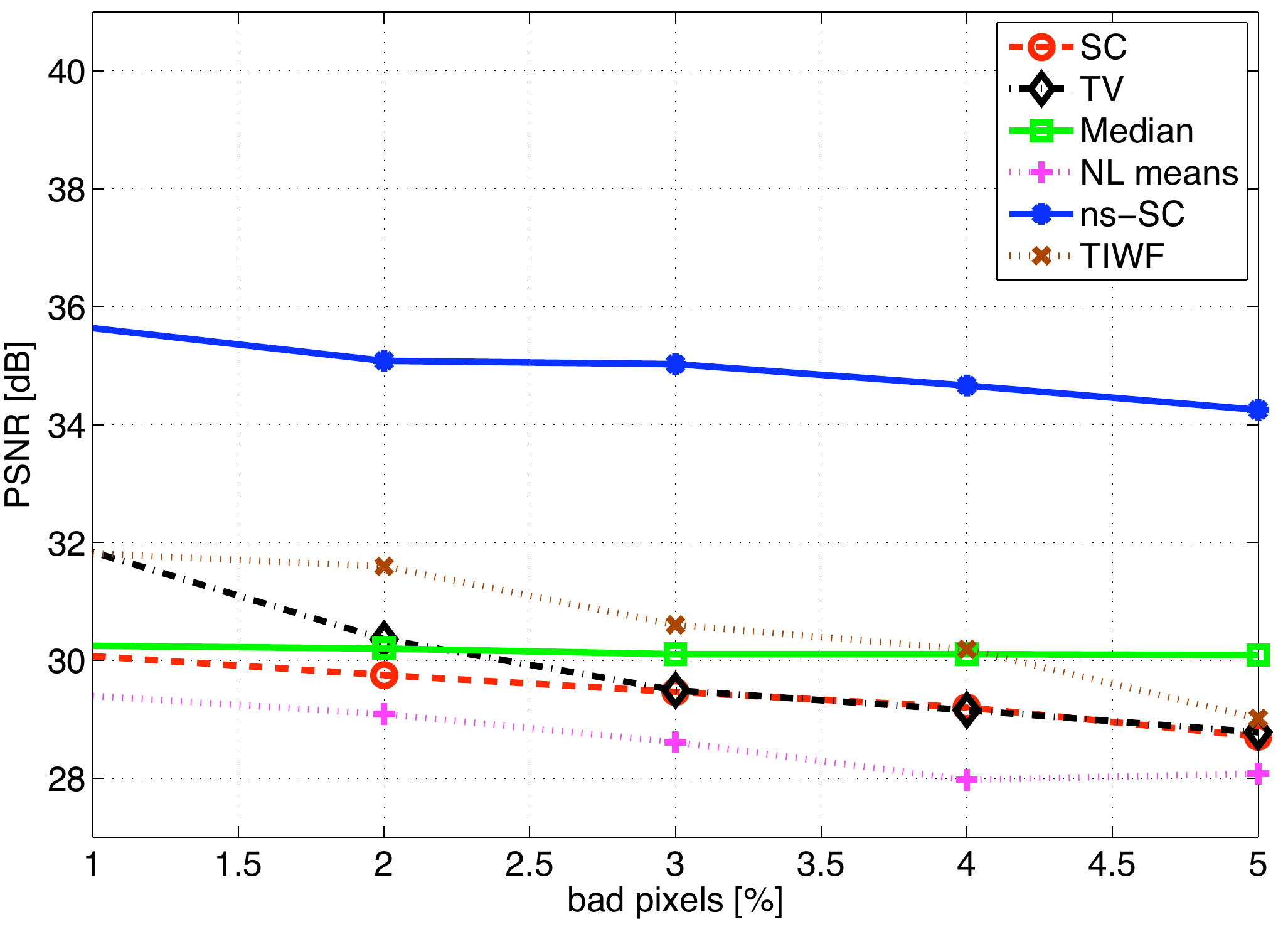}\\
\mbox{(a) } & \mbox{(b) } \\
\end{array}$
\caption{Average denoising performance with synthetic noise: PSNR vs the \% of corrupted pixels (averaged over five depth maps from the Purves dataset). a) Denoising performance comparison of ns-SC using different dictionaries $\mathbf{\Phi}_1$, $\mathbf{\Phi}_2$ and $\mathbf{\Phi}_3$. b) Denoising performance comparison of ns-SC (using $\mathbf{\Phi}_1$) with other denoising methods: SC - sparse coding with $\mathbf{\Phi}_3$; TV - total variation with optimal $\lambda$; Median filtering; NL means - non-local means; TIWF - non-stationary sparse inference with the translation invariant wavelet 7-9 frame.}
\label{fig:psnr}
\end{center}
\end{figure*}

Fig.~\ref{fig:psnr}a and Fig.~\ref{fig:psnr}b show PSNR versus the \% of corrupted pixels averaged over five depth maps, and it confirms the superiority of ns-SC with $\mathbf{\Phi}_1$ over ns-SC with other dictionaries (Fig.~\ref{fig:psnr}a) and over other denoising solutions (Fig.~\ref{fig:psnr}b). As expected, ns-SC using a dictionary learned without masking the missing pixels ($\mathbf{\Phi}_2$, Approach 2) performs worse than ns-SC using a dictionary learned with masking ($\mathbf{\Phi}_1$, Approach 1), since less information is provided to the learning algorithm\footnote{Note that the denoising algorithm does not use pixel masking, the masking refers only to the way that the dictionary is learned.}. Interestingly, ns-SC with $\mathbf{\Phi}_2$ performs better than SC with $\mathbf{\Phi}_1$, which shows the superiority of ns-SC compared to SC. Another advantage of ns-SC compared to other denoising methods is that we do not need to choose a special value for the regularization parameter $\lambda$ since the noise variance at each pixel is inferred during denoising. On the other hand, for TV denoising and NL means we have chosen the values of $\lambda$ or values of the noise variance that yield the best results.

\begin{figure*}[!htbp]
\begin{center}
$\begin{array}{c@{\hspace{0.1 cm}}c@{\hspace{0.1 cm}}c@{\hspace{0.1 cm}}c@{\hspace{0.1 cm}}c@{\hspace{0.1 cm}}c}
\includegraphics[width=0.19\textwidth]{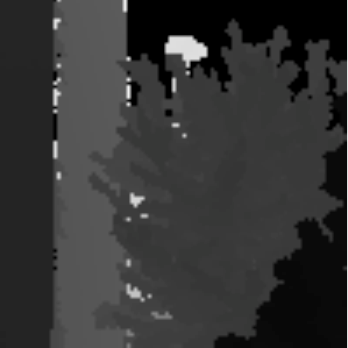}&
\includegraphics[width=0.19\textwidth]{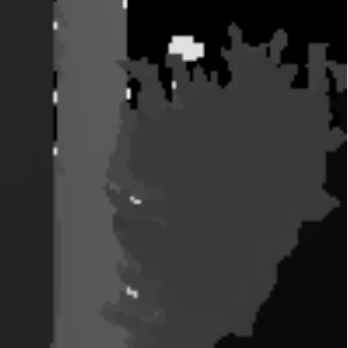}&
\includegraphics[width=0.19\textwidth]{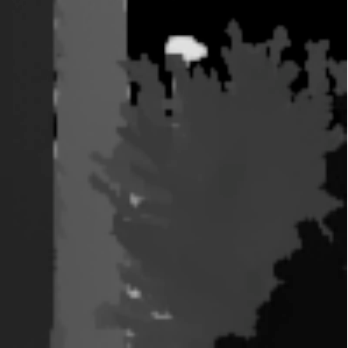}&
\includegraphics[width=0.19\textwidth]{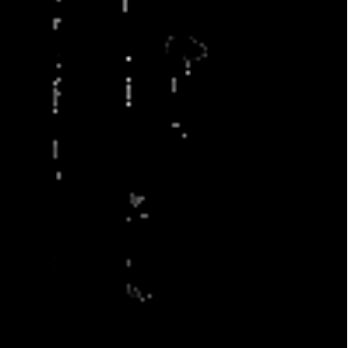} &
\includegraphics[width=0.19\textwidth]{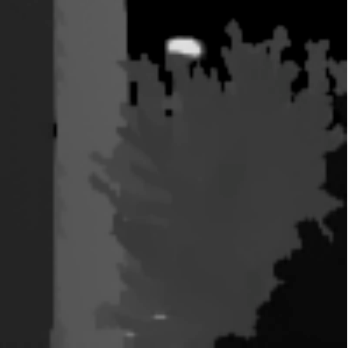}\\
\mbox{ \footnotesize{(a) original}} & \mbox{ \footnotesize{(b) NL-means} } & \mbox{ \footnotesize{(c) ns-SC }}& \mbox{ \footnotesize{(d) variance from ns-SC}}& \mbox{ \footnotesize{(e) inpainted ns-SC}}
\end{array}$
\end{center}
\caption{Denoising results for a depth map from the Purves database with natural noise. Noisy samples appear mostly around depth edges, but there is also some correlated noise (big white region). NL means filtering (b) and ns-SC (c) both remove only uncorrelated noise, but the NL means filter also undesirably smoothens the textured regions. In addition to denoising, ns-SC infers the variance of the noise (d), which points out the noisy samples (non-zero pixels). The indication of noisy samples can be used for inpainting (e).}
\label{fig:denoisingpurves2}
\end{figure*}

If we have a depth map that has natural noise due to the acquisition process, we can use ns-SC to remove the noise, but also to point out the noisy pixels. These would correspond to the pixels whose inferred noise variance is high. Fig.~\ref{fig:denoisingpurves2}a shows a noisy laser range scan depth map from the Purves database, where the noisy pixels take values within the depth range (i.e., they are not marked as missing pixels). Reconstructions of  this depth map obtained by NL means filtering and ns-SC (with $\mathbf{\Phi}_1$) are shown in Fig.~\ref{fig:denoisingpurves2}b and Fig.~\ref{fig:denoisingpurves2}c, respectively. Both methods denoised most of the erroneous pixels, but the NL-means filtering also introduced a loss of texture information, while ns-SC preserves this information. Moreover, ns-SC gives an estimate of the noise variance at each pixel, displayed in Fig.~\ref{fig:denoisingpurves2}d. Interestingly, the indication of noisy pixels can be used as a mask for inpainting by setting those pixel variances to infinity (due to the finite precision of ns-SC, the inferred variances are finite). The final inpainted depth map with ns-SC is given in Fig.~\ref{fig:denoisingpurves2}e.

\begin{figure*}[!htbp]
\begin{center}
$\begin{array}{c@{\hspace{0.2 cm}}c@{\hspace{0.2 cm}}c@{\hspace{0.2 cm}}c@{\hspace{0.2 cm}}c}
\includegraphics[width=0.22\textwidth]{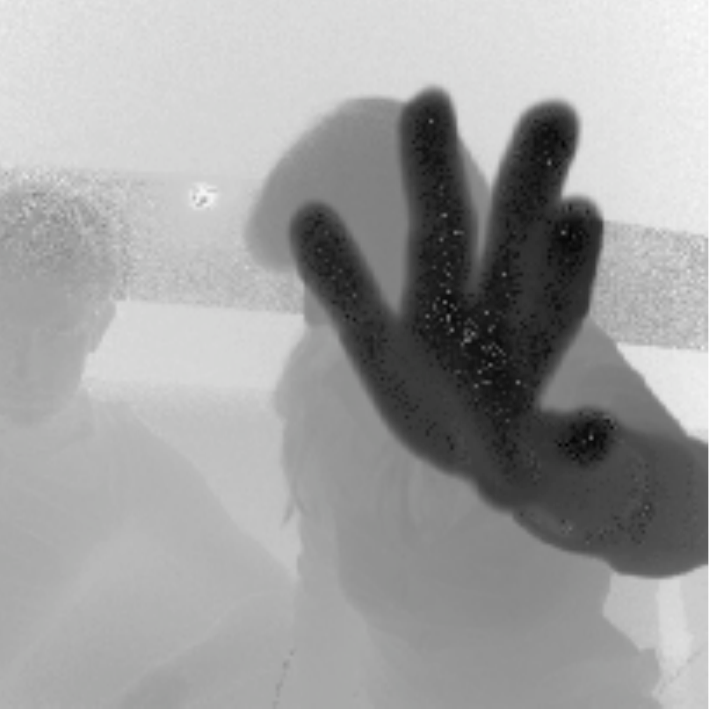}&
\includegraphics[width=0.22\textwidth]{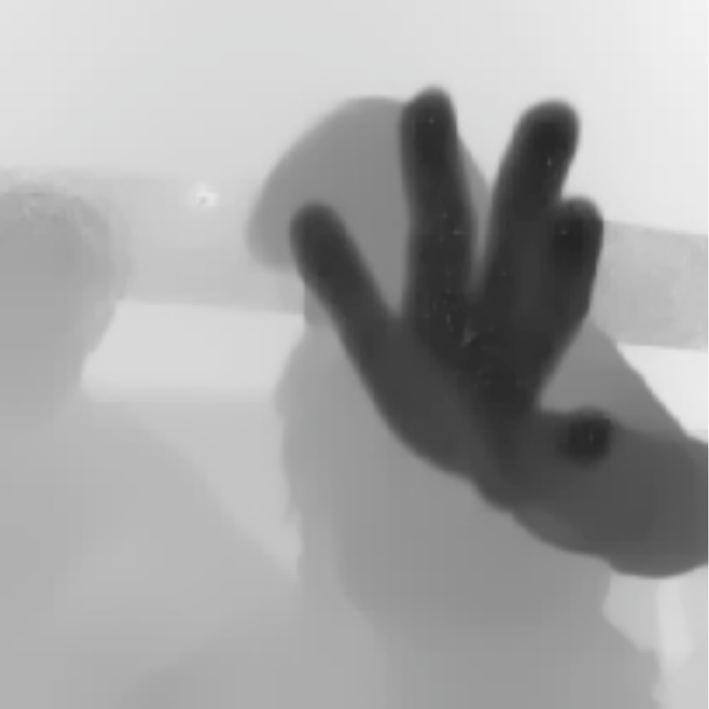}&
\includegraphics[width=0.22\textwidth]{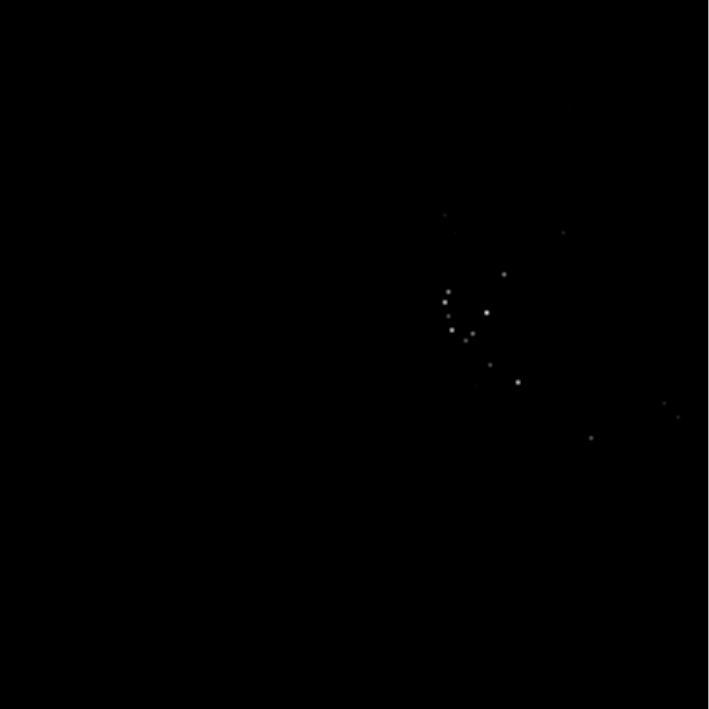}&
\includegraphics[width=0.22\textwidth]{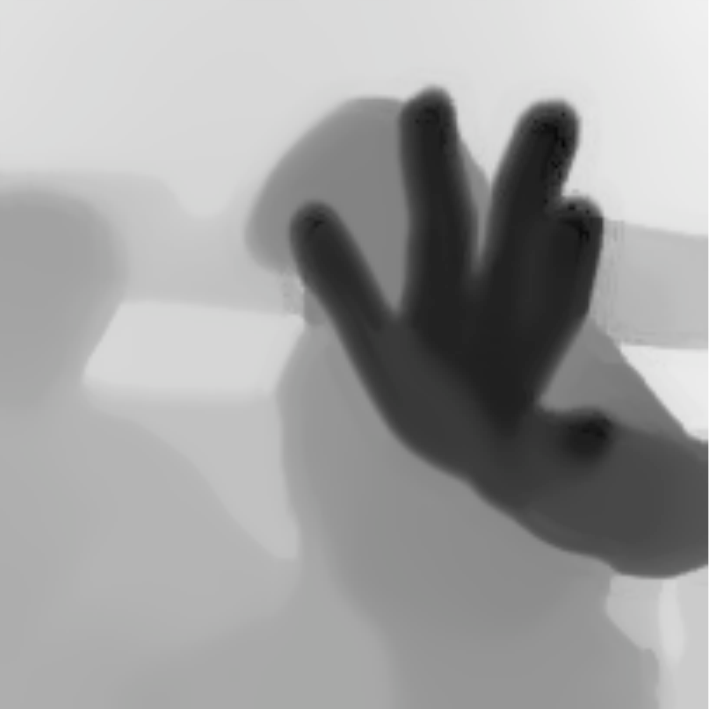} \\
\mbox{ (a) original} & \mbox{(b) denoised ns-SC } & \mbox{(c) variance from ns-SC }& \mbox{(d) denoised with NLmeans}
\end{array}$
\end{center}
\caption{Denoising results for a depth map from the time of flight depth image with the measurement noise. NL means filtering (d) denoises the depth map, but smoothens out the fine details. In addition to good denoising, ns-SC (b) infers the variance of the noise (d), which points out the noisy samples (non-zero pixels).}
\label{fig:denoisingTOF}
\end{figure*}

Similar results are obtained on the depth maps captured by the PMD Time-Of-Flight (TOF) camera\footnote{\url{http://www.pmdtec.com/}}. These cameras capture 2D video + depth dynamic information with a reasonable time resolution. The camera sends a modulated optical signal to the environment and measures the time of the round trip travel of light for each pixel. The result is a depth map, which is usually very noisy. The original and denoised depth maps are given in Fig.~\ref{fig:denoisingTOF}a-d. Compared to NL-means, ns-SC preserves some fine detail while removing the noise. 

\subsection{Stereo matching}
Finally, we show the results of the two-layer stereo matching algorithm described in Section~\ref{sec:stereo} using two different algorithms for solving the MRF in the middle layer: 1) the graph cut (GC) algorithm~\cite{Boykov:2001p4748}, and 2) the second order prior (2OP) algorithm~\cite{Woodford09}. Although these are not the top performing algorithms on the Middlebury benchmark, they are among the most widely known, and their code is available online. Since our two-layer model does not require a specific optimization algorithm for the MRF (e.g., graph cut), an interested reader can apply an algorithm of choice. Here, we are primarily interested in evaluating the performance improvement obtained by adding a second layer of hidden units above the MRF solved with the two mentioned exemplary algorithms.

We have used a dictionary learned with Approach 1 (Sec.~\ref{subsec:results_learning}), with $32\times32$ pixel atoms, and dictionary size of 1024 atoms. The learned dictionary has atoms similar to $\mathbf{\Phi}_1$, but larger, which leads to increased inference efficiency for larger depth maps.

\begin{figure*}[!htbp]
\begin{center}
$\begin{array}{c@{\hspace{0.05 cm}}c@{\hspace{0.05 cm}}c@{\hspace{0.05 cm}}c@{\hspace{0.05 cm}}c}
\includegraphics[width=0.195\textwidth]{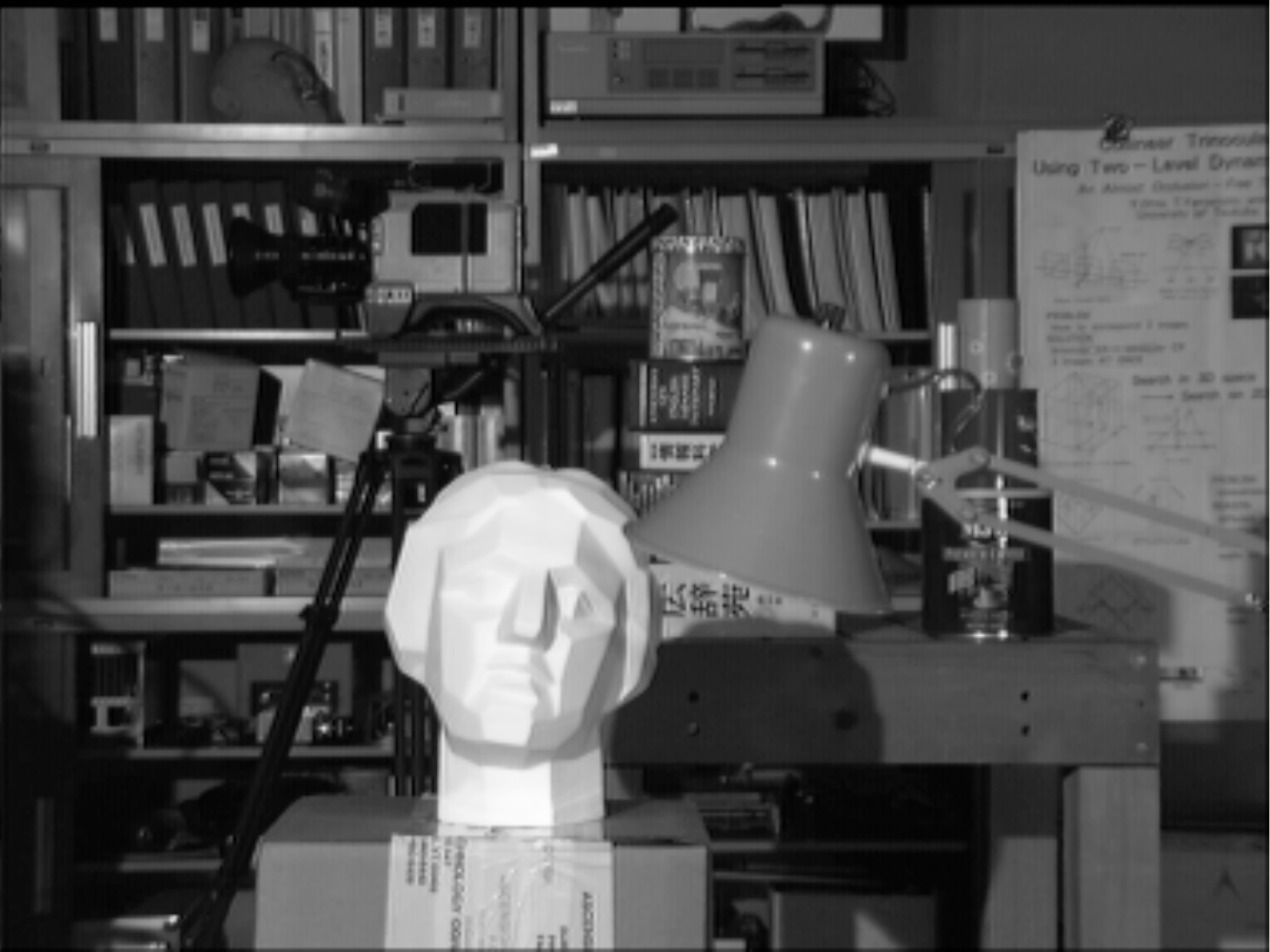} &
\includegraphics[width=0.195\textwidth]{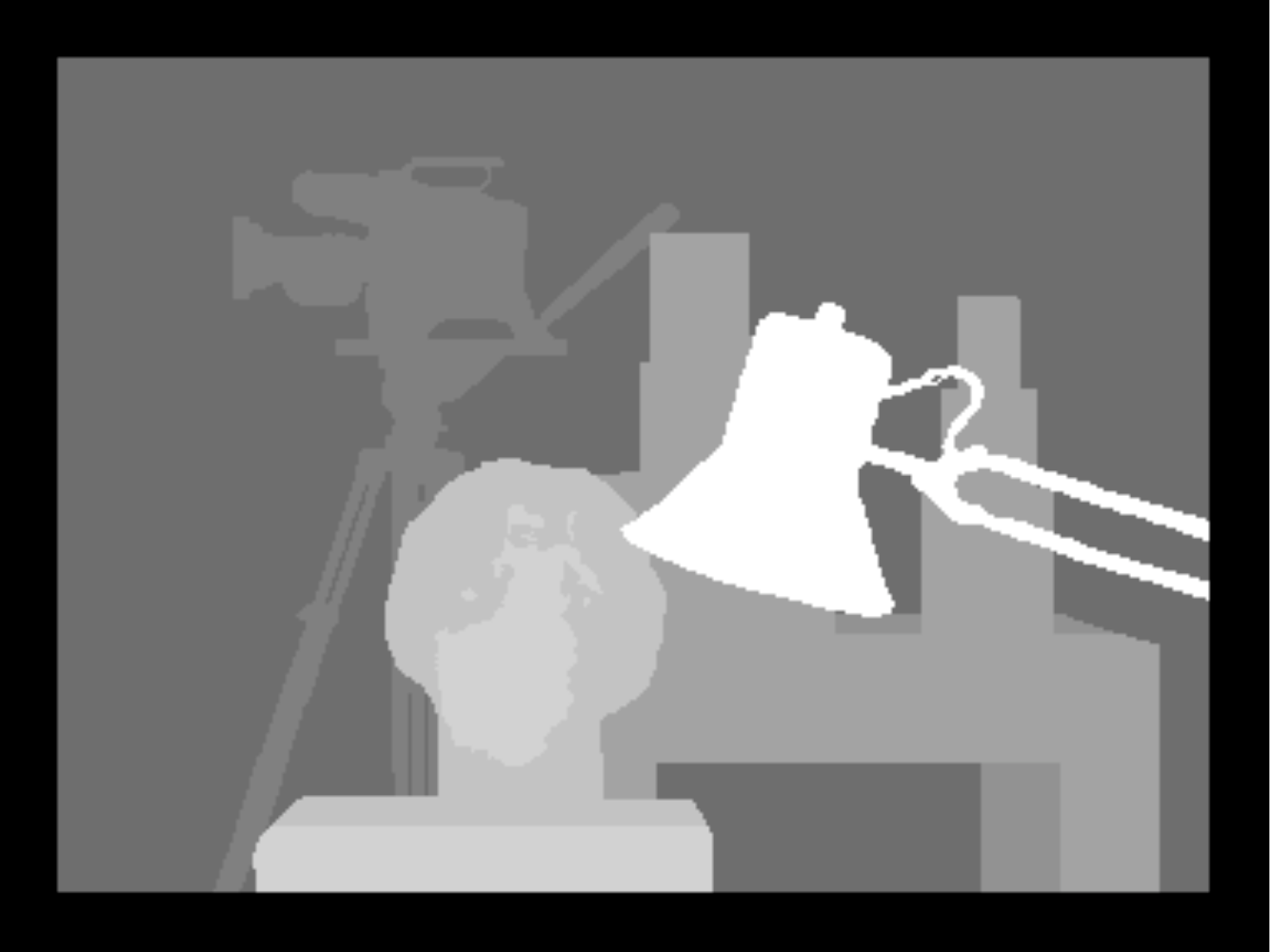}&
\includegraphics[width=0.195\textwidth]{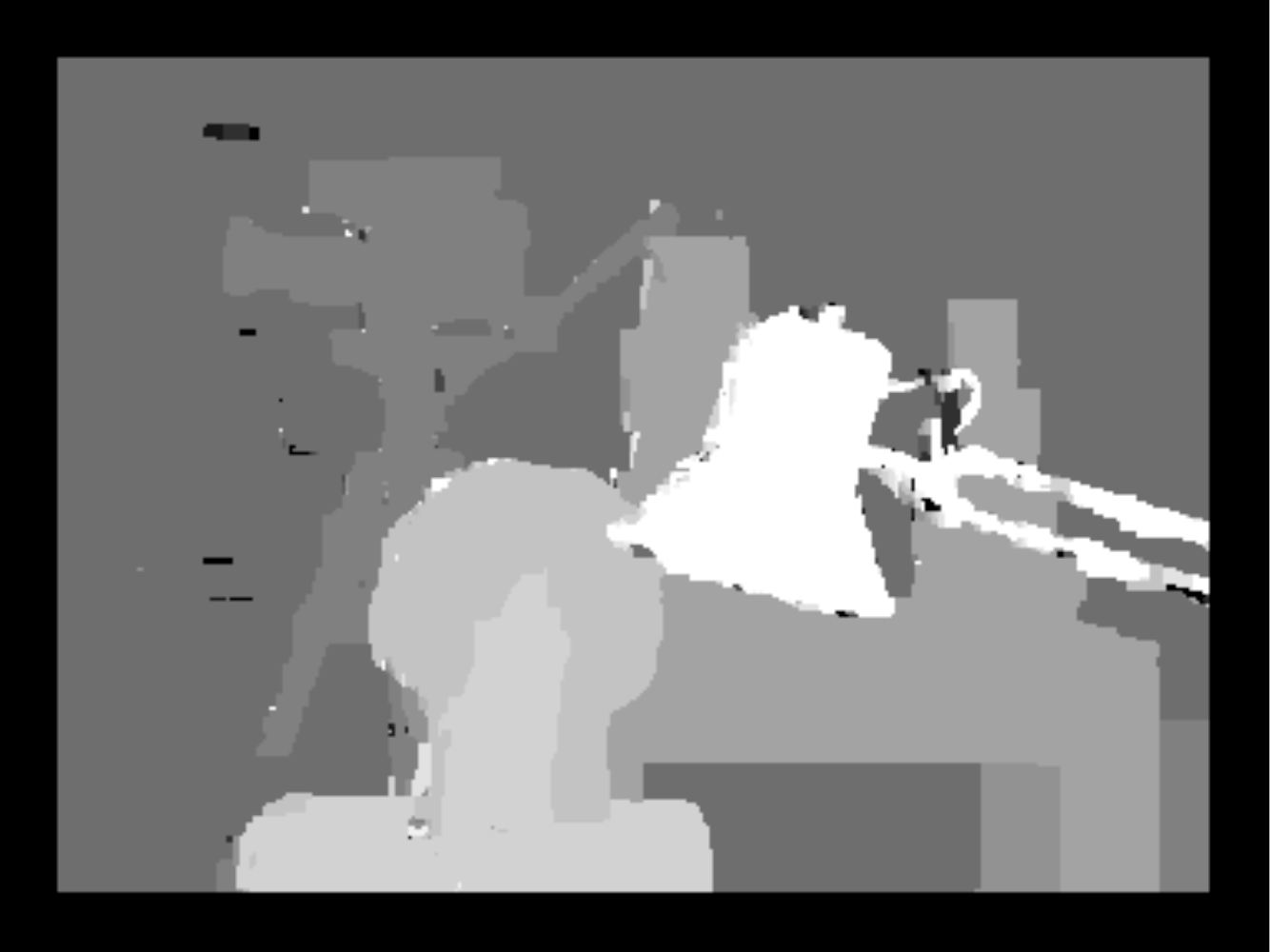} &
\includegraphics[width=0.195\textwidth]{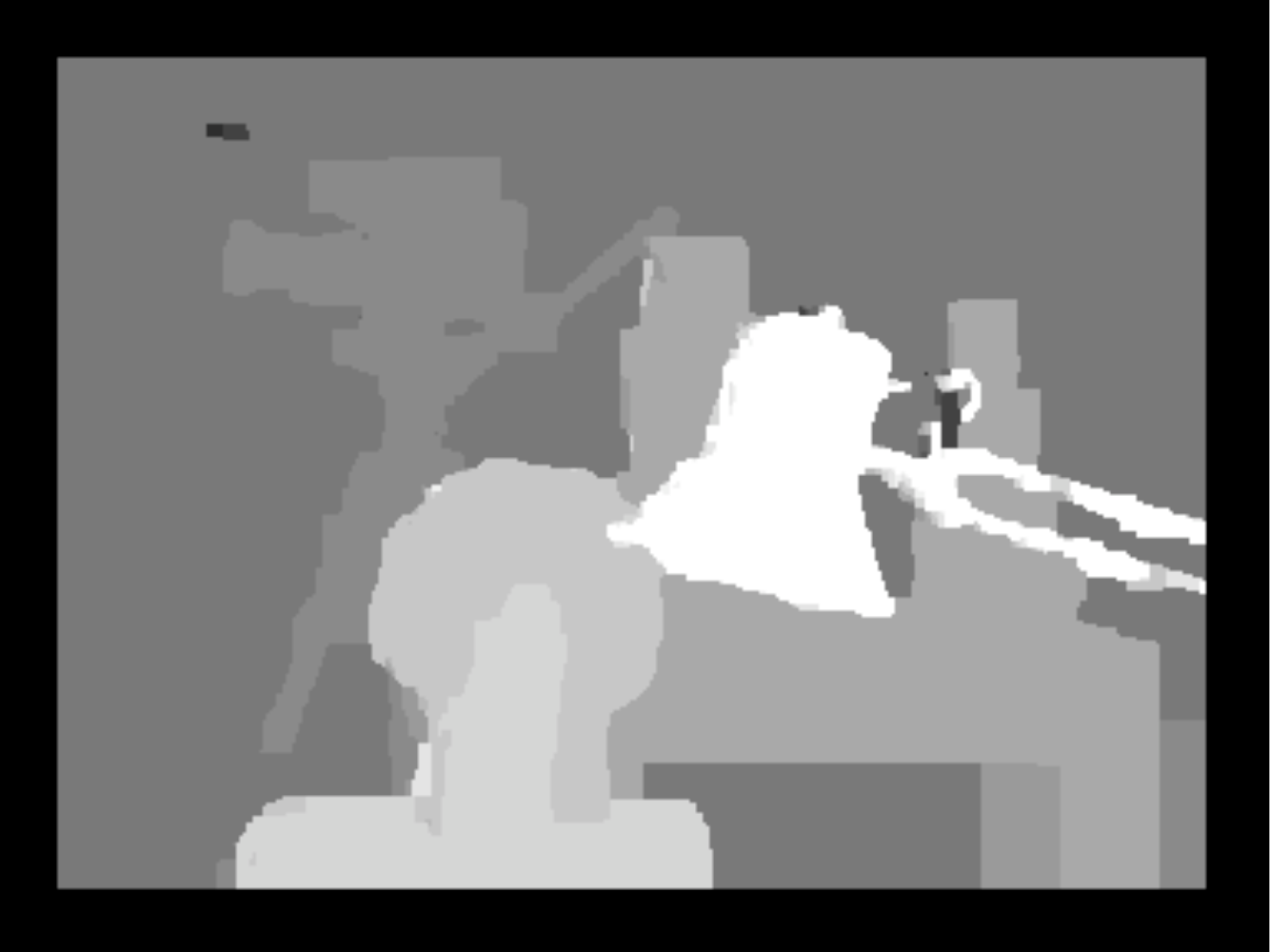}  &
\includegraphics[width=0.195\textwidth]{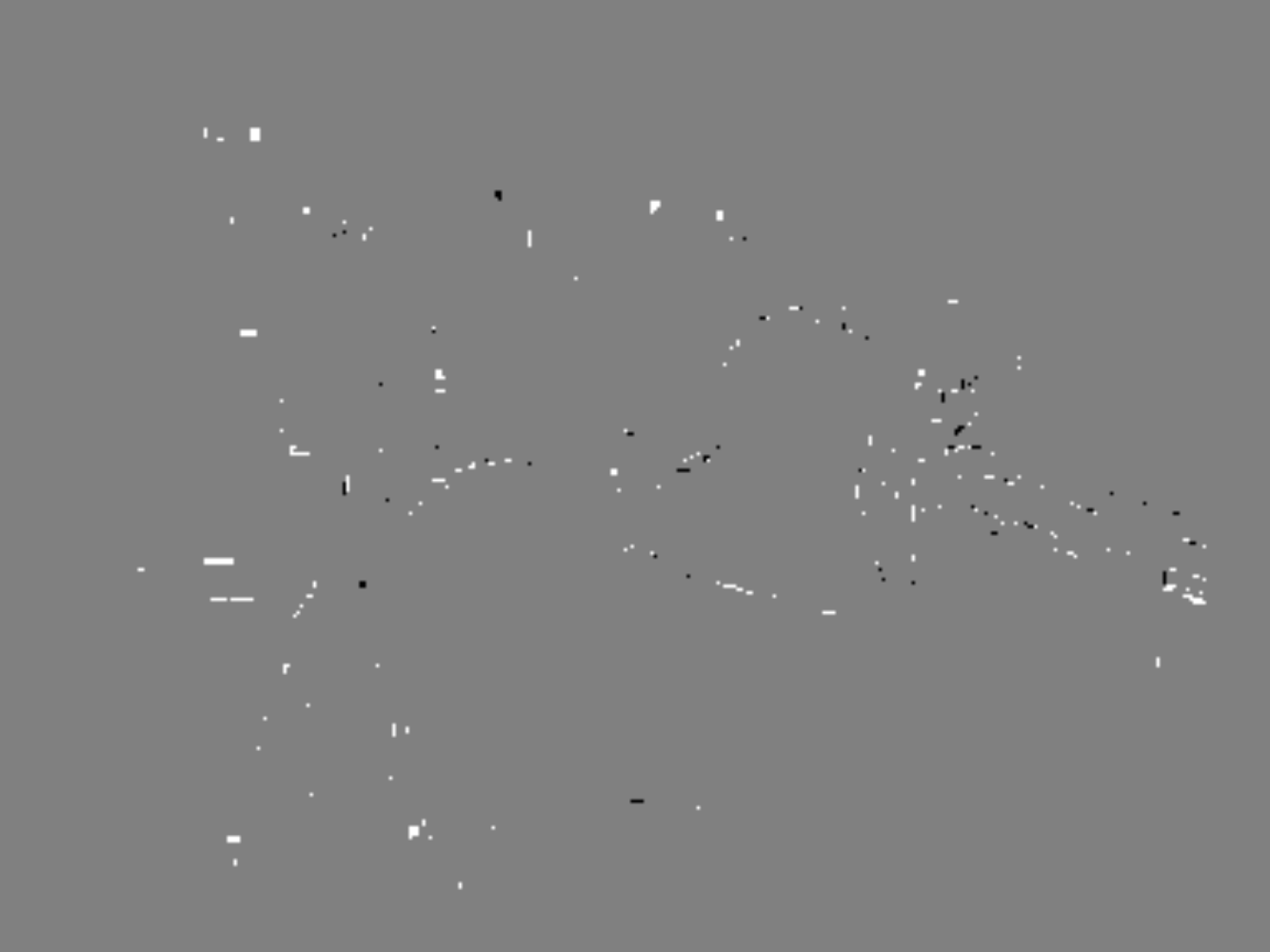} \\[0.05cm]
\footnotesize\mbox{ (a) left image} & \footnotesize\mbox{(b) ground truth} & \footnotesize\mbox{ (c) GC (9.58\%)} & \footnotesize\mbox{(d) GC + ns-SC (9.21\%)} & \footnotesize\mbox{ (e) modified pixels}
\end{array}$
\end{center}
\caption{Disparity estimation results on the Tsukuba dataset: a) left image $288\times 384$); b) ground truth disparity map; c) disparity estimation result with graph cut~\cite{Boykov:2001p4748}, percentage of bad pixels:  9.58\%, d) disparity estimation result with the two layer (GC+ns-SC) model with the learned dictionary (patch size $32 \times 32$), percentage of bad pixels:  9.21\%, e) modified pixels: white - correctly modified by the upper layer, black - falsely modified by the upper layer.}
\label{fig:tsukuba}
\end{figure*}

Fig.~\ref{fig:tsukuba} shows the performance of GC and GC + ns-SC (our two layer model with GC in the middle layer) on the Tsukuba stereo pair from the 2001 Middlebury stereo database, which does not belong to the training set. The original left image from the stereo pair is shown in Fig.~\ref{fig:tsukuba}a (the right image is similar), while the ground truth disparity map is given in Fig.~\ref{fig:tsukuba}b. The estimated disparity map using the alpha-beta swap graph cut algorithm~\cite{Boykov:2001p4748,1316848,Kolmogorov04}, is shown in Fig.~\ref{fig:tsukuba}c. It uses the data term with equal variances $\rho_i^2$ for single cliques, and the Potts energy model for pairwise cliques, given as: $V_2(f_i,f_j)=u_{\{i,j\}} T(f_i\neq f_j)$, where $u_{\{i,j\}}=U(|L_i-R_j|)$ and $T$ denotes the indicator function. The function $U$ is defined as:
\[ U(|L_i-R_j|) = \left\{ \begin{array}{ll}
         \mbox{$2K$} & \mbox{if $|L_i-R_j| < 5$} \, ;\\
         \mbox{$K$} & \mbox{otherwise}.\end{array} \right. \]
The parameter $K$ has been set to 20, which is in the range proposed in~\cite{Boykov:2001p4748}. Fig.~\ref{fig:tsukuba}d shows the estimated disparity map using our two layer model. This map was obtained at convergence after three iterations between the disparity inference in two layers. Inference in the bottom layer is done using the graph cut with the modified data term and single cliques according to Eq.~(\ref{eq:depthEns}). Since ns-SC assumes uncorrelated noise, it infers the noise variance in single pixels of the disparity map. Therefore, the pairwise cliques are not affected. Even though graph cut returns discrete estimates of the disparity, this does not change the continuous nature of the optimization in the upper layer, as the quantization error is subsumed in $(f_i - \hat{f}_i)^2$. We can see that the two layer model improves the graph cut result by correcting 0.37\% of pixels. The obtained disparity map is also visually improved. Fig.~\ref{fig:tsukuba}e shows the map of pixels modified by adding the upper layer of sparse nodes: white pixels denote the correctly modified pixels (from erroneous to correct) and black denote falsely corrected pixels (from correct to erroneous). Clearly, there are more correctly modified pixels, which are mostly located around depth edges. This is consistent with the fact that the learned dictionary contains oriented edges.
%\footnote{Note that one can get a smoother disparity map by changing the parameters of the graph cut, but that does not reduce the error as it wrongly connects separate 3D objects.}

\begin{figure*}[!htbp]
\begin{center}
$\begin{array}{c@{\hspace{0.05 cm}}c@{\hspace{0.05 cm}}c@{\hspace{0.05 cm}}c@{\hspace{0.05 cm}}c}
\includegraphics[width=0.195\textwidth]{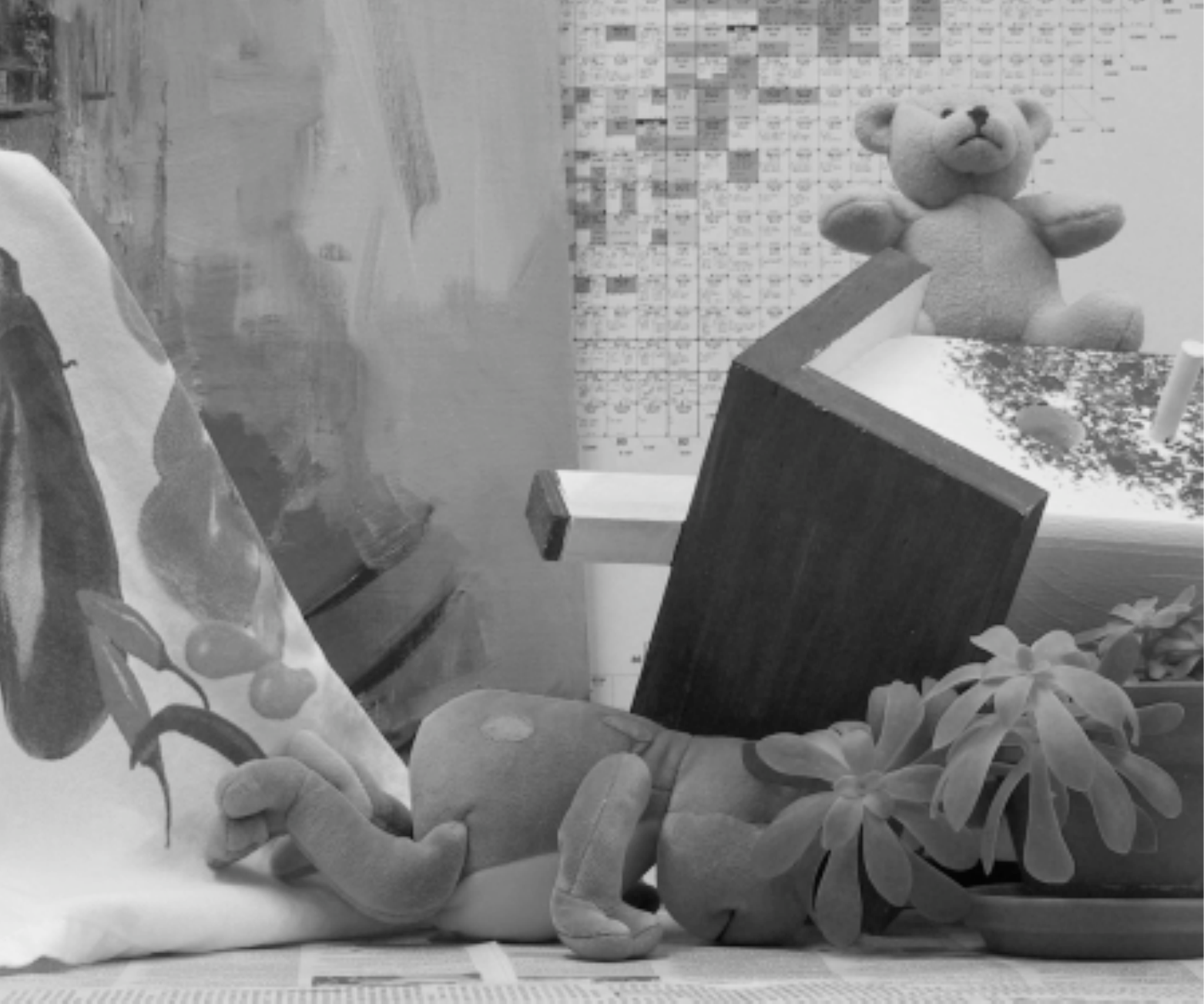} &
\includegraphics[width=0.195\textwidth]{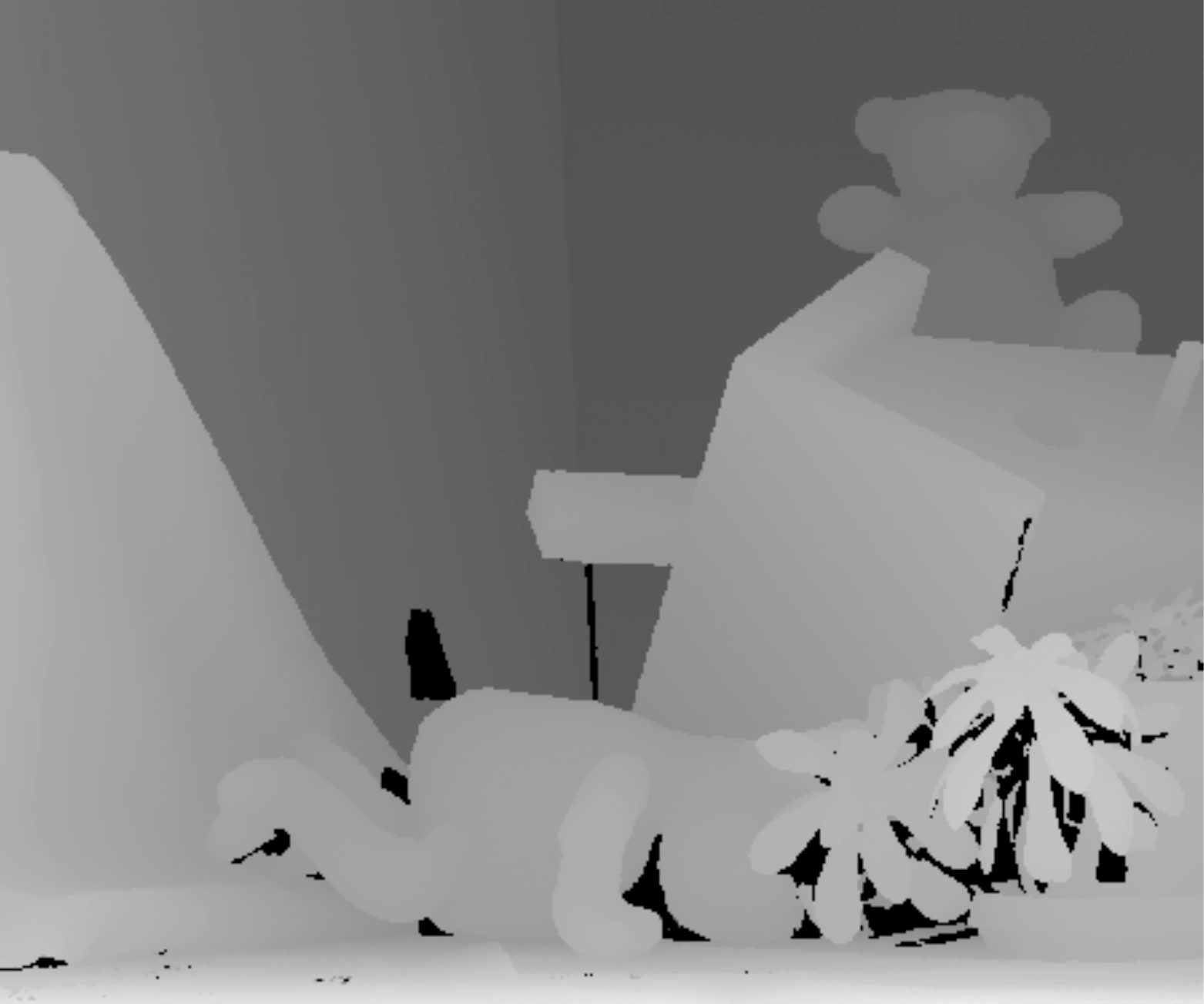} &
\includegraphics[width=0.195\textwidth]{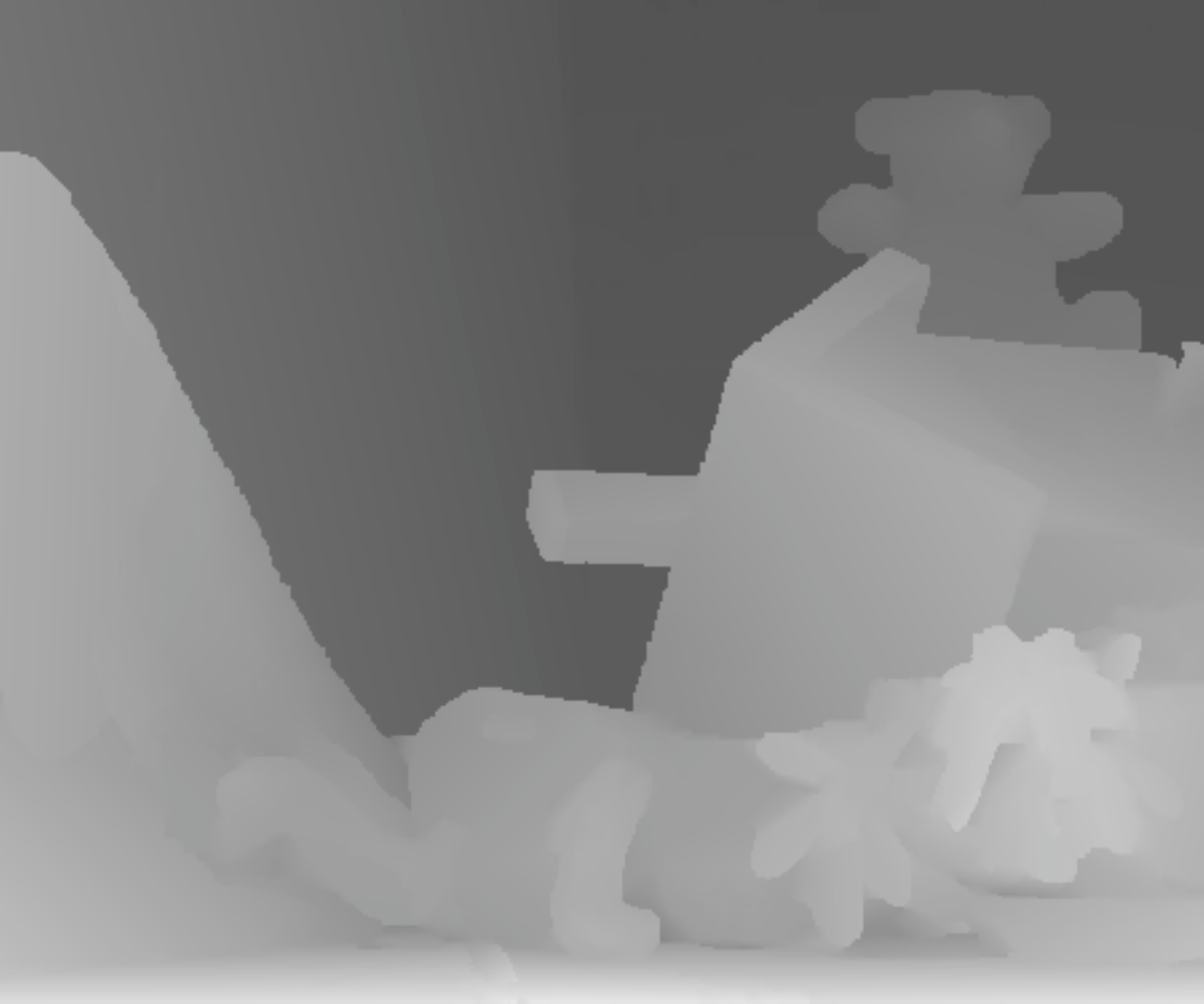} &
\includegraphics[width=0.195\textwidth]{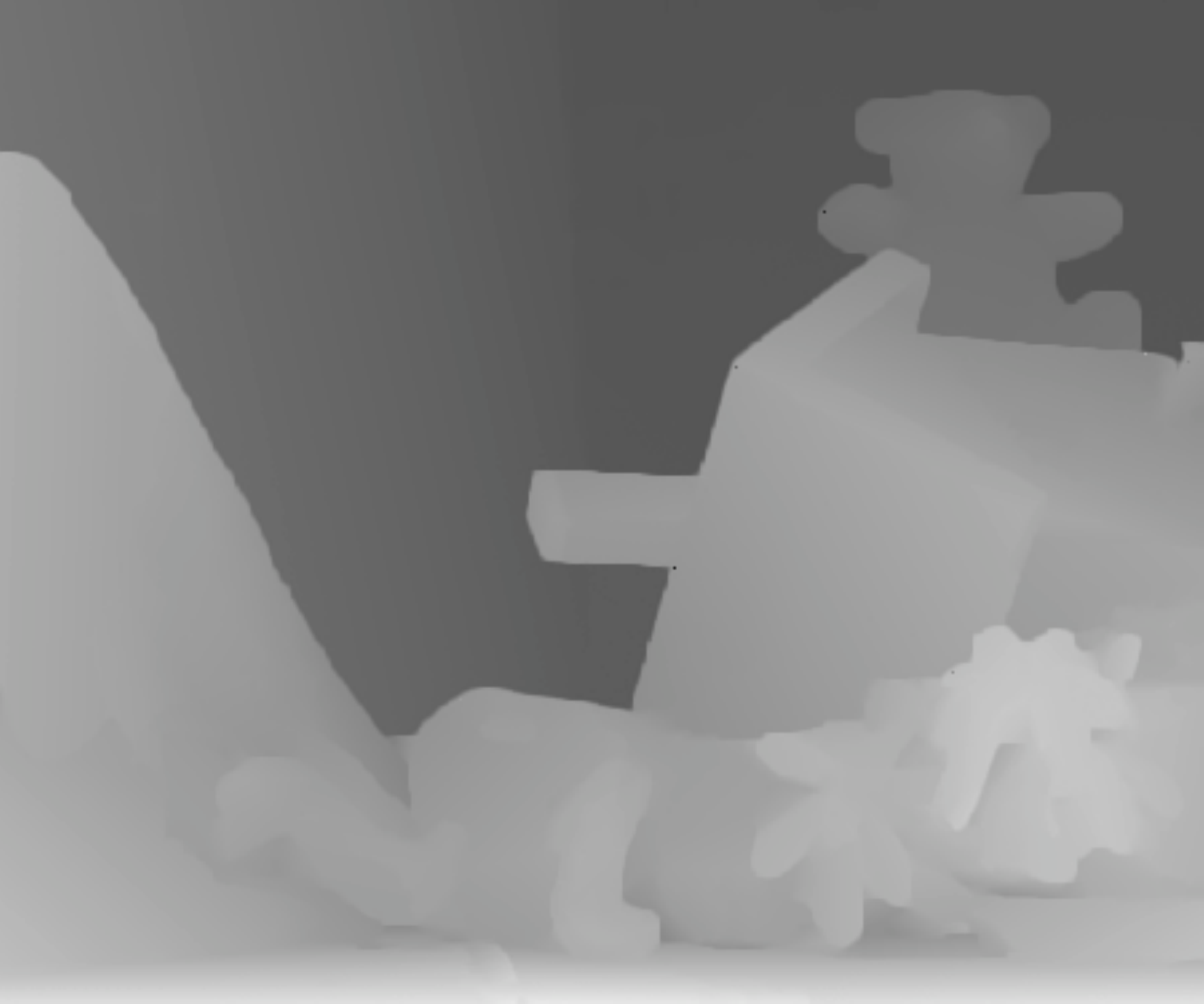}&
\includegraphics[width=0.195\textwidth]{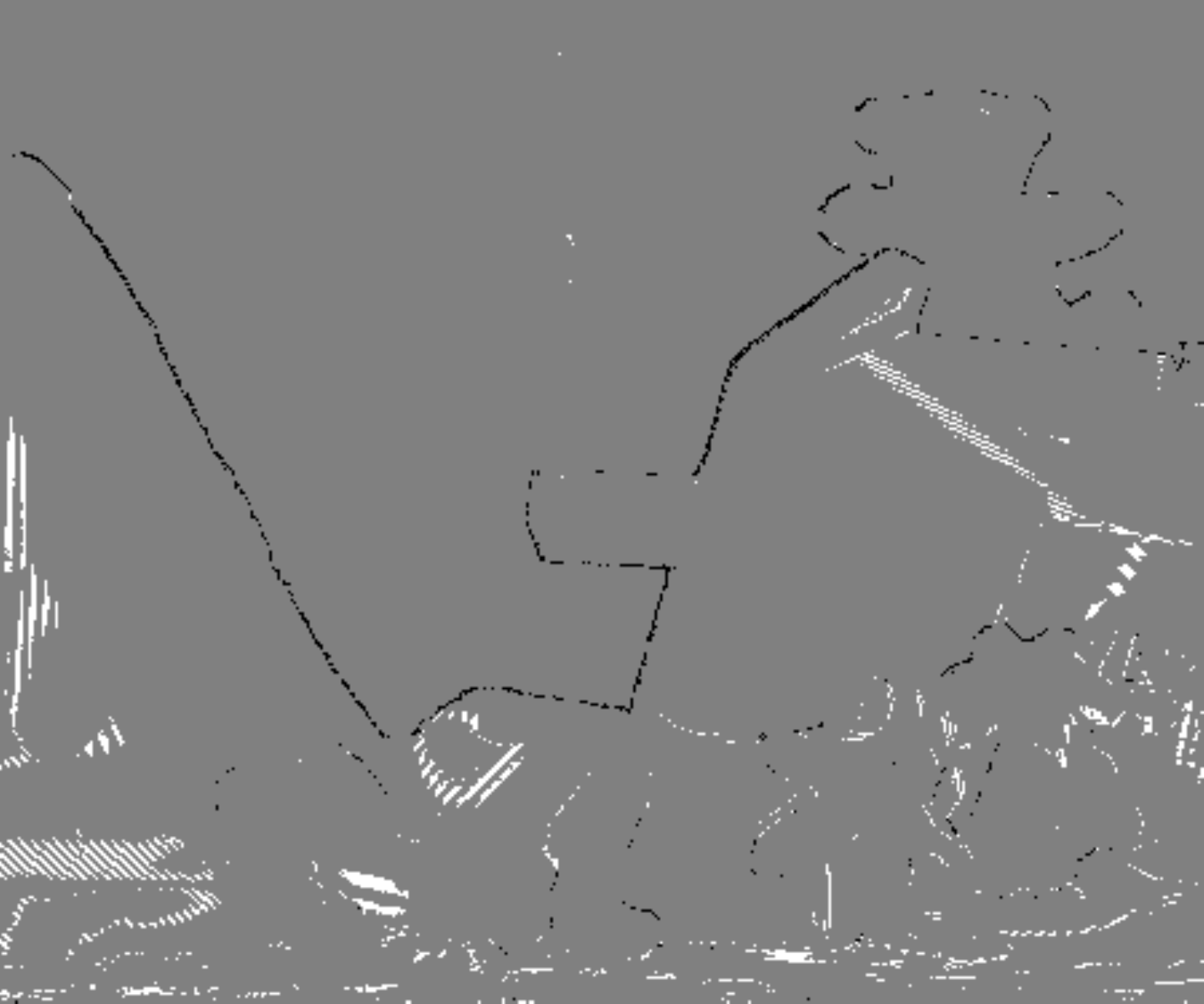}  \\[0.05cm]
\footnotesize\mbox{ (a) left image} & \footnotesize\mbox{ (b) ground truth} & \footnotesize\mbox{(c) 2OP (9.40\%)} & \footnotesize\mbox{ (d) 2OP+ns-SC (8.14\%)} & \footnotesize\mbox{ (e) modified pixels}
\end{array}$
\end{center}
\caption{Disparity estimation results on the Teddy dataset: a) left image ($375\times 450$); b) ground truth disparity map; c) disparity estimation result with the 2OP algorithm~\cite{Woodford09}, percentage of bad pixels:  9.40\%, d) disparity estimation result with the two layer (2OP+ns-SC) model with the learned dictionary (patch size $32 \times 32$), percentage of bad pixels:  8.14\%, e) pixels: white - correctly modified by the upper layer, black - falsely modified by the upper layer.}
\label{fig:teddy}
\end{figure*}

\begin{figure*}[!htbp]
\begin{center}
$\begin{array}{c@{\hspace{0.05 cm}}c@{\hspace{0.05 cm}}c@{\hspace{0.05 cm}}c@{\hspace{0.05 cm}}c}
\includegraphics[width=0.195\textwidth]{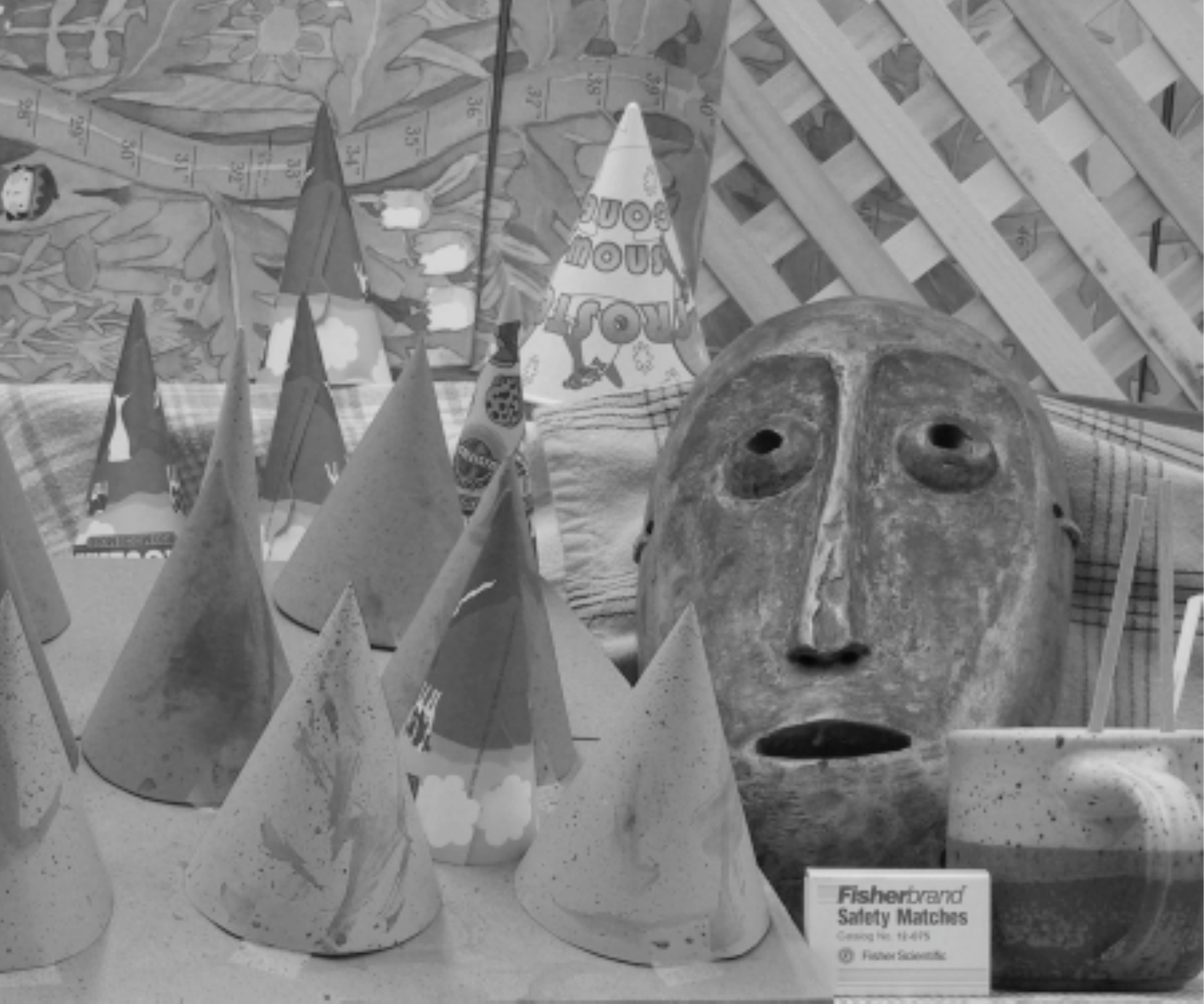} &
\includegraphics[width=0.195\textwidth]{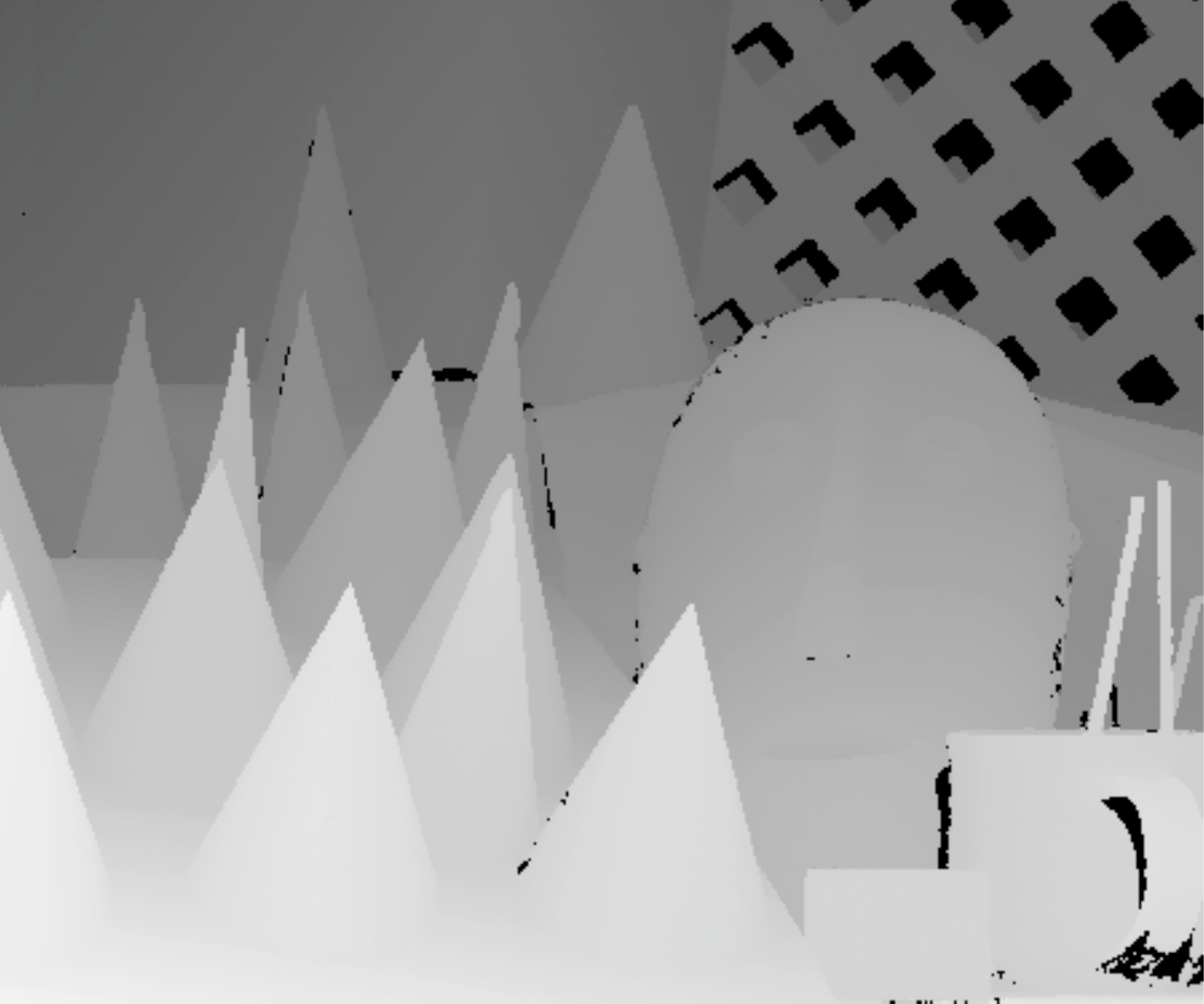} &
\includegraphics[width=0.195\textwidth]{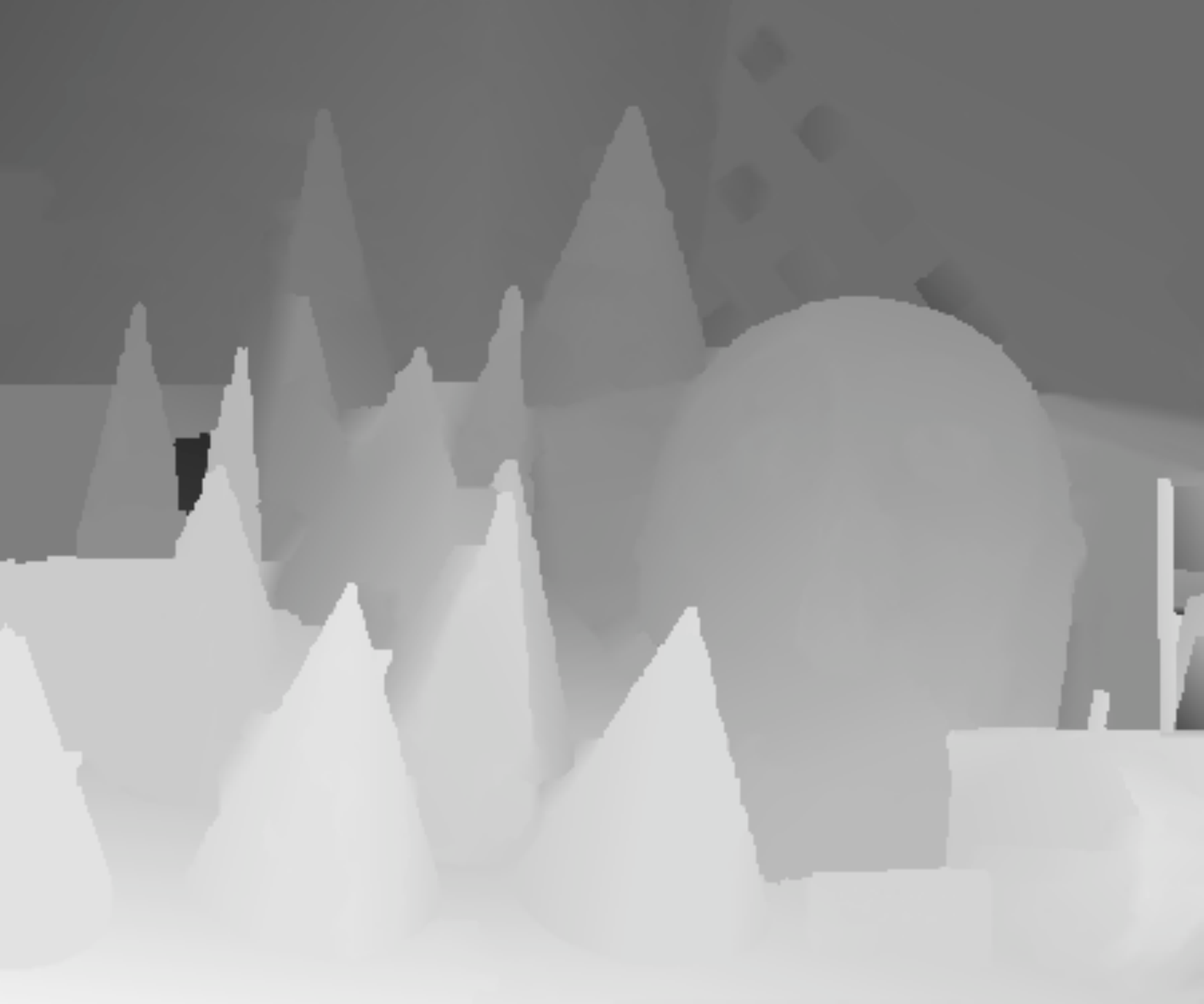} &
\includegraphics[width=0.195\textwidth]{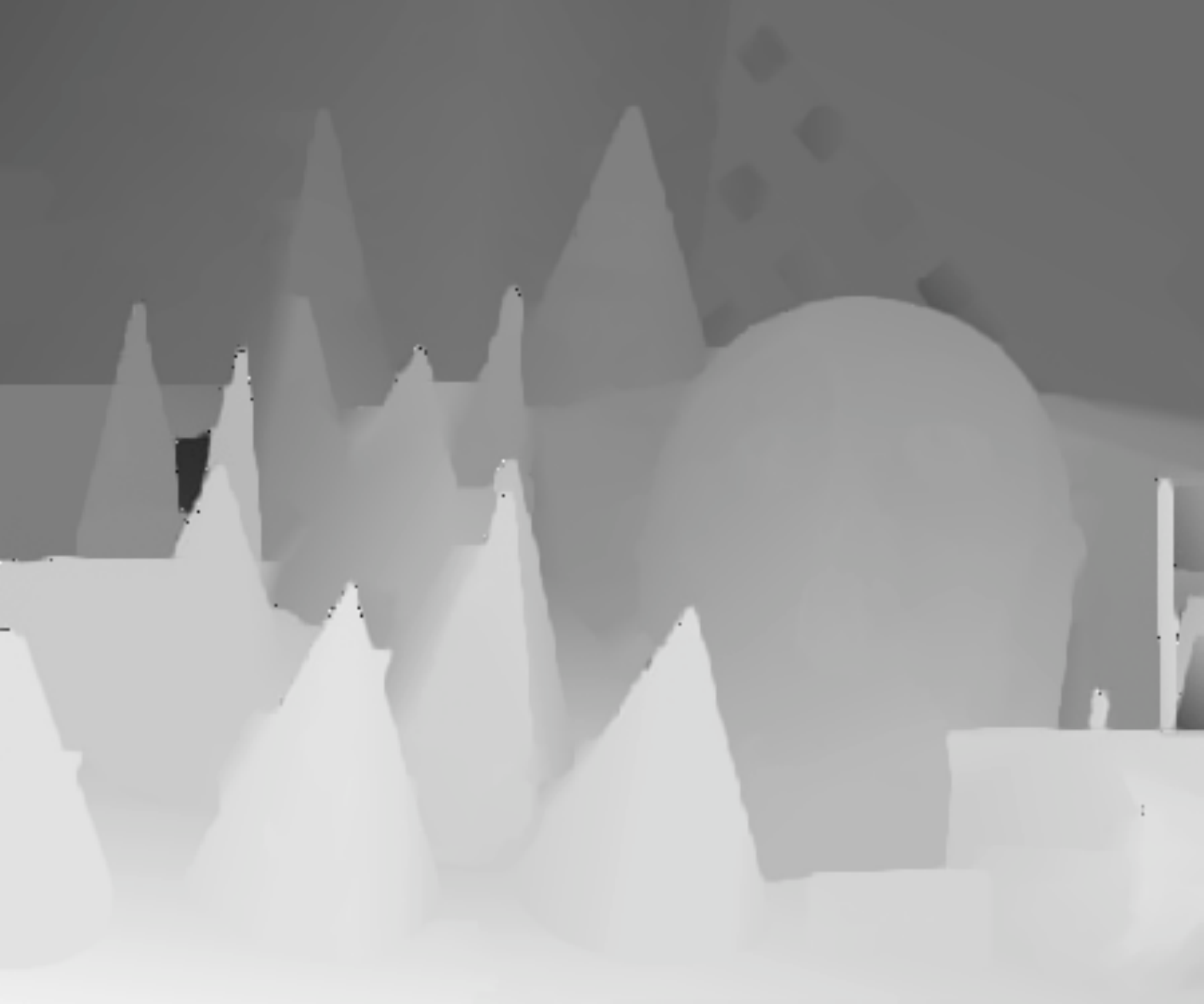}&
\includegraphics[width=0.195\textwidth]{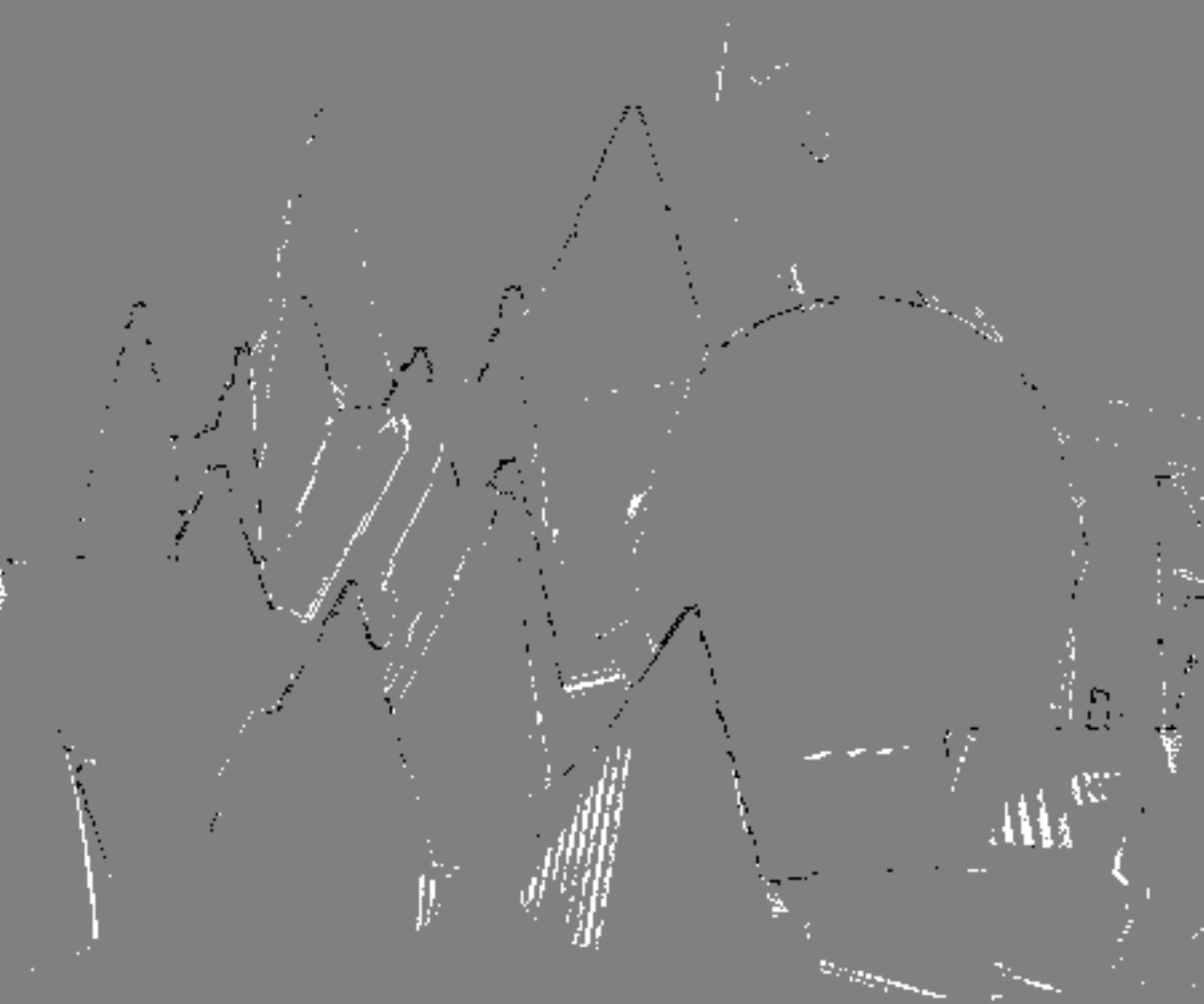}  \\[0.05cm]
\footnotesize\mbox{ (a) left image} & \footnotesize\mbox{(b) ground truth} & \footnotesize\mbox{ (c) 2OP (12.73\%)} & \footnotesize\mbox{ (d) 2OP+ns-SC (11.98\%)} & \footnotesize\mbox{ (e) modified pixels}
\end{array}$
\end{center}
\caption{Disparity estimation results on the Cones dataset: a) left image ($375\times 450$); b) ground truth disparity map; c) disparity estimation result with the 2OP algorithm~\cite{Woodford09}, percentage of bad pixels:  12.73\%, d) disparity estimation result with the two layer (2OP+ns-SC) model with the learned dictionary (patch size $32 \times 32$), percentage of bad pixels:  11.98\%, e) modified pixels: white - correctly modified by the upper layer, black - falsely modified by the upper layer.}
\label{fig:Cones}
\end{figure*}

To demonstrate the generality of the two-layer method, we have also tested it using the 2OP algorithm of Woodford et al.~\cite{Woodford09}, which solves the MRF with triplewise cliques. The code and the default parameters have been obtained from the authors website\footnote{\url{http://www.robots.ox.ac.uk/~ojw/2op/index.html}}. After only two iterations between the two layers in our model, the percentage of bad pixels has been reduced with respect to the 2OP algorithm by 1.26\% on the Teddy dataset (Fig.~\ref{fig:teddy}), and by 0.75\% on the Cones dataset (Fig.~\ref{fig:Cones}), where the disparity accuracy is set to one pixel. The error corresponding to missing pixels (black pixels on the ground truth map) has not been taken into account. Both datasets are from the Middlebury database and do not belong to the training set. Since the 2OP algorithm does random initialization, we have ran the inference five times, and obtained the average improvement of 1.25\% for the Teddy set and 0.92\% for the Cones set. In each run, the two layer model consistently outperformed the 2OP algorithm. Interestingly, the upper layer correctly modified pixels that are mostly located on the surfaces and some around edges, but also falsely modified some depth edges (see Fig.~\ref{fig:teddy}e and Fig.~\ref{fig:Cones}e). False modification of some depth edges is due to the fact that the image segmentation strategy of the 2OP algorithm leads to better estimation of depth around boundaries, compared to just using the depth priors given by the upper layer. This underlines the importance of learning the joint statistics of depth and intensity, which represents a promising direction of future research.

\section{Related work}

\subsection{Depth map representation and denoising}
Although learning representations of images has been widely addressed in the literature in the last few decades~\cite{olshausen97sparse, Engan:focus,Lewicki:2000p2757,Aharon:2006p284, mairal:214,Yaghoobi09,5332363}, there has not been much work on learning representations of depth/disparity maps. The most closely related work to the one presented in this paper is the work by Mahmoudi and Sapiro, who learn sparse representations of depth maps for surface reconstruction~\cite{Mahmoudi09}. Their work differs from ours in two aspects: 1) they assume a stationary Gaussian noise model; and 2) they learn overcomplete dictionaries per shape, i.e., the dictionaries do not generalize to a set of depth maps. As we have seen in Section~\ref{sec:results}, depth dictionary learning with stationary additive noise yields inferior denoising performance compared to the one with a non-stationary noise model. Besides the advantage of learning under adaptation to noise/unreliabilty of depth maps, our method also infers the measure of unreliability of each pixel in a depth map, which is extremely useful in applications such as view synthesis~\cite{view} and stereo matching~\ref{subsec:stereo}.

In the context of depth coding/compression, some researchers have proposed constructions of transforms that are adapted to shape and depth representation. Maitre and Do proposed a shape-adaptive wavelet transform that generates a small number of wavelet coefficients along depth edges~\cite{Maitre2010513}. The coding scheme allocates more bits for representing depth edges, which are detected by the Canny edge detector. Construction of piecewise smooth functions, called ``platelets" represents also an interesting approach for dealing with smooth images with sharp boundaries, such as confocal microscopy images~\cite{1199635} or depth maps~\cite{morvan:60550K}. Methods~\cite{Maitre2010513} and~\cite{morvan:60550K} demonstrate efficient coding performance on ground truth (not noisy) depth maps. To the best of our knowledge, these methods have not been extended to deal with noisy, uncertain depth maps usually obtained from stereo matching or time of flight cameras.

In addition to the problem of dealing with unreliable depth estimates in image based rendering, denoising of depth maps has become of significant interest recently due to the development of the Time-Of-Flight (TOF) cameras. Unlike in standard imaging, the noise in depth maps is non-stationary: it has different statistics for different scene contents. Interestingly, the noise variance of depth pixels is inversely proportional to the amplitude values of light captured by the sensor pixels~\cite{Edeler09}. Edeler et al. used this relation and a non-stationary noise model equivalent to ours in order to perform superresolution of depth maps~\cite{Edeler09}. However, their solution of the inverse problem assumes known noise statistics particular to TOF data, while our approach infers these statistics and thus it is more general. The aforementioned relation between noise variances and light amplitudes does not hold for the noise introduced around depth edges and at close distances, when it becomes more of ``salt and pepper" nature, i.e., there are depth outliers. Most previous work on TOF depth data denoising deals with this noise type by first removing the outliers and then denoising the depth map using the bilinear filter~\cite{KCLU07} or non-local (NL) means~\cite{Huhle08} (also see~\cite{Schall2008701} for the application of NL-means to laser range data). Prior removal of outliers is crucial here, since these would bias the estimate of the noise variance for the depth map. Our work does not need outlier removal, since those are inferred along the sparse approximation algorithm. Moreover, we obtain a quantitative estimate of the reliability for each pixel in the depth map. The experimental results in Section~\ref{sec:results} confirm that our approach is superior to NL means filtering using the median filter, which is better suited for salt-and-pepper noise and does not smooth out the discontinuities in the depth map. Finally, one should note that the proposed variance inference represents a general way of estimating the noise statistics and it can be used in many regularization-based framework for denoising (e.g., in a variational formulation of NL means). Such variance-adaptive denoising strategy would certainly improve the performance of those methods on depth data. It thus represents an important contribution to the field of denoising.

\subsection{Depth from stereo}
As it is a highly ill-posed problem, stereo correspondence significantly depends on prior information about the depth structure in the scene. The most significant progress in stereo matching has been made by utilizing the depth smoothness prior. Although the idea that nearby pixels should have similar disparities dates back to the seventies~\cite{Marr76}, high performance depth estimation algorithms appeared much later with the introduction of the piecewise smoothness ($l_1$) prior that preserves depth discontinuities. Depth estimation algorithms, such as the graph cut~\cite{Boykov:2001p4748} and belief propagation~\cite{JianSun:2003p290}, define the matching problem as an energy minimization problem, where the depth map is modeled as a Markov Random Field (MRF)~\cite{MRF} with single and pairwise clique potentials. Due to the significant performance improvements with respect to previous approaches, these MRF-based algorithms gained a lot of success. During the last decade, many methods based on graph cuts and belief propagation have been proposed, which attain performance improvements by including additional constraints to handle occlusions~\cite{GCocc,BPocc,BPocc2} or by performing image segmentation during or prior to matching~\cite{BPseg}.

Other modifications of MRF approaches include algorithms that extend the MRF objective from single and pairwise cliques to triplewise~\cite{Woodford09} and higher order ones~\cite{Ishikawa09}. Since graph-cut algorithms cannot be straightforwardly extended to optimally solve MRFs that include priors on these higher order dependencies~\cite{Kolmogorov04}, methods~\cite{Woodford09} and~\cite{Ishikawa09} are based on the QPBO~\cite{4270228} optimization algorithm. However, QPBO gives suboptimal solutions for higher order priors, leaving a certain number of pixels unlabeled in estimated disparity maps. Moreover, the computational complexity of QPBO increases exponentially with the degree of the MRF, and limiting implementations to triple-wise cliques.

Since depth maps of natural scenes contain more complex structures that cannot by captured by pairwise or triple-wise statistics, it is important to include priors on higher order dependencies in stereo matching. The proposed two layer approach to depth inference offers an efficient way to regularize the solution of stereo matching by utilizing sparse priors over learned depth dictionaries. Because of its generic formulation, the proposed method can use any of the state of the art MRF-based depth estimation methods in the middle layer, and obtain an improved depth map solution. The most important contribution of the proposed two-layer approach to stereo matching is that it can be applied so generally.

\section{Conclusions}

We have presented a method to learn dictionaries of depth features that capture higher-order dependencies in depth maps, resulting in oriented depth edges and slanted surfaces. Because depth is not a perfectly measurable phenomenon, learning its statistics has to be performed under noisy conditions, where the type of noise is significantly different than the one usually seen in images. Our new sparse coding algorithm explicitly takes into account the noisy nature of depth estimates, such that the inference and learning can ``see through" the noise in order to fill in and learn the appropriate structure. Moreover, it infers a reliability measure of each sample that can be further used in any algorithm having inferred data as input. Our denoising results have demonstrated that the depth dictionary learned with the new ns-SC method with non-stationary noise gives superior performance compared to the state of the art. 

The sparsity prior that enforces higher order dependencies is then exploited in a new stereo matching method. We have defined a two-layer graphical model where the nodes in the middle layer encode disparities and their correlation, and the nodes in the upper layer enforce sparse priors. The proposed approach is quite general: the inference in the middle layer can use any existing MRF-based depth estimation algorithm, which combined with sparse inference in the upper layer can yield improved performance. The importance of higher order dependencies in the depth structure is confirmed by the superior performance of the two layer model compared to the MRF-based model only, for two different MRF-based algorithms. A promising perspective is to use ns-SC to learn joint representations of texture (color) and depth. It will also be important to go beyond linear generative models to properly deal with occlusion in 3D scenes.

\section{Acknowledgements}
The authors would like to thank Dr. Zafer Arican from the Signal Processing Laboratory LTS4, EPFL, Switzerland, for his help in recording the video data captured by a time of flight camera.

\small
\bibliographystyle{IEEEtran}
\bibliography{stereo-nips}

%References follow the acknowledgments. Use unnumbered third level heading for
%the references. Any choice of citation style is acceptable as long as you are
%consistent. It is permissible to reduce the font size to `small' (9-point) 
%when listing the references. {\bf Remember that this year you can use
%a ninth page as long as it contains \emph{only} cited references.}

\end{document}